\documentclass[preprint]{elsarticle}

\usepackage[utf8]{inputenc}

\usepackage{amssymb}

\usepackage{tikz}

\usepackage[utf8]{inputenc}

\usepackage{mathrsfs} 
\usepackage{amssymb,amsmath,amsthm,graphicx,float}
\usepackage{mathtools}
\usepackage{booktabs}
\usepackage{tabularx}
\newcommand*{\arial}{\fontfamily{phv}\selectfont}
\usepackage{caption}
\usepackage{subcaption}

\usepackage{algorithm}
\usepackage{algpseudocode}

\usepackage[toc,page]{appendix}

\newtheorem{problem}{Problem}

\newtheorem{remark}{Remark}

\DeclareMathOperator*{\argmax}{arg\,max}

\newcommand{\Pomdp}{\mathcal{M}}

\newcommand{\States}{\mathcal{S}}
\newcommand{\state}{s}
\newcommand{\action}{\alpha}
\newcommand{\Actions}{\mathcal{A}}
\newcommand{\Transition}{\mathcal{P}}
\newcommand{\Observations}{\mathcal{O}}
\newcommand{\ObservationSet}{\mathcal{Z}}
\newcommand{\observation}{z}

\newcommand{\StateOccup}{\mu}
\newcommand{\StateActionOccup}{\nu}

\newcommand{\Reward}{\mathcal{R}}
\newcommand{\Initdist}{\StateOccup_0}
\newcommand{\discount}{\gamma}
\newcommand{\Real}{\mathbb{R}}

\newcommand{\Distr}{\mathrm{Distr}}
\newcommand{\policy}{\sigma}
\newcommand{\Rewardvec}{\theta}
\newcommand{\trustregion}{\rho}
\newcommand{\slack}{k}
\newcommand{\mapping}{\eta}
\newcommand{\memoryupt}{\delta}
\newcommand{\target}{\mathcal{T}}

\newcommand{\reachPropSymbol}{\varphi}
\usepackage{tikz,pgfplots}
\pgfplotsset{compat=1.8}
\usepackage{pgfplotstable}
\usepgfplotslibrary{groupplots}
\usepackage[framemethod=default]{mdframed}

\usepackage{times}
\usepackage{bbm}
\usepgfplotslibrary{fillbetween}
\usetikzlibrary{arrows,decorations.pathmorphing,positioning,fit,trees,shapes,shadows,automata,calc} 
\usetikzlibrary{patterns,arrows,arrows.meta,calc,shapes,shadows,decorations.pathmorphing,decorations.pathreplacing,automata,shapes.multipart,positioning,shapes.geometric,fit,circuits,trees,shapes.gates.logic.US,fit}
\usetikzlibrary{backgrounds,scopes}
\usepackage{aligned-overset}

 \newenvironment{customlegend}[1][]{%
        \begingroup
        \csname pgfplots@init@cleared@structures\endcsname
        \pgfplotsset{#1}%
    }{%
        \csname pgfplots@createlegend\endcsname
        \endgroup
    }%

    \def\addlegendimage{\csname pgfplots@addlegendimage\endcsname}
 \newcommand\titlesize{\fontsize{8.1pt}{10.2pt}\selectfont}

\usepackage{newfloat}
\usepackage{listings}
\floatstyle{ruled}
\newfloat{listing}{tb}{lst}{}
\floatname{listing}{Listing}

\newcommand{\myett}[1]{#1}

\begin{document}

\begin{frontmatter}


\title{\myett{Task-Guided IRL in POMDPs that Scales}} 

\author{Franck Djeumou\corref{cor1}}
\ead{fdjeumou@utexas.edu}
\cortext[cor1]{The University of Texas at Austin}

\author{Christian Ellis\corref{cor2}}
\ead{cellis3@umassd.edu}
\cortext[cor2]{The University of Massachusetts: Dartmouth}

\author{Murat Cubuktepe\corref{cor1}}
\ead{mcubuktepe@utexas.edu}

\author{Craig Lennon\corref{cor3}}
\ead{ craig.t.lennon.civ@army.mil}
\cortext[cor3]{United States Army Research Laboratory}

\author{Ufuk Topcu\corref{cor1}}
\ead{utopcu@utexas.edu}




\begin{abstract}
In inverse reinforcement learning (IRL), a learning agent infers a reward function encoding the underlying task using demonstrations from experts.
However, many existing IRL techniques make the often unrealistic assumption that the agent has access to full information about the environment.
We remove this assumption by developing an algorithm for IRL in partially observable Markov decision processes (POMDPs).
We address two limitations of existing IRL techniques. First, they require an excessive amount of data due to the information asymmetry between the expert and the learner. 
Second, most of these IRL techniques require solving the computationally intractable \emph{forward problem}\textemdash computing an optimal policy given a reward function\textemdash in POMDPs. The developed algorithm reduces the information asymmetry while increasing the data efficiency by incorporating task specifications expressed in temporal logic into IRL. Such specifications may be interpreted as side information available to the learner a priori in addition to the demonstrations. Further, the algorithm avoids a common source of algorithmic complexity by building on causal entropy as the measure of the likelihood of the demonstrations as opposed to entropy. Nevertheless, the resulting problem is nonconvex due to the so-called \emph{forward problem}. We solve the intrinsic nonconvexity of the forward problem in a scalable manner through a sequential linear programming scheme that guarantees to converge to a locally optimal policy. In a series of examples, including experiments in a high-fidelity Unity simulator, we demonstrate that even with a limited amount of data and POMDPs with tens of thousands of states, our algorithm learns reward functions and policies that satisfy the task while inducing similar behavior to the expert by leveraging the provided side information.
\end{abstract}



\end{frontmatter}

\section{Introduction}
A robot can satisfy certain human-specified tasks by describing desired behavior through a reward function.
However, the design of such a reward function is a non-trivial task.
Inverse reinforcement learning (IRL) is an established technique that infers a reward function encoding the underlying task using expert demonstrations.
IRL techniques have found a wide range of applications in various domains such as acrobatic helicopter flight~\cite{abbeel2010autonomous}, inferring future actions of people~\cite{kitani2012activity}, human-autonomy interaction~\cite{hadfield2016cooperative,dragan2013policy}, robotic surgery~\cite{osa2014online,osa2017online}, and robotic manipulation tasks~\cite{finn2016guided}.
Most existing work~\cite{abbeel2010autonomous,ziebart2008maximum,zhou2017infinite,ziebart2010modeling,hadfield2016cooperative,finn2016guided} has focused on Markov decision processes (MDPs), assuming that the learner can fully observe the state of the environment and expert demonstrations.
However, the learner will not have access to full state observations in many applications.
For example, a robot will never know everything about its environment~\cite{ong2009pomdps,bai2014integrated,zhang2017robot} and may not observe the internal states of a human with whom it works~\cite{akash2019improving,liu2012modeling}.
Such information limitations violate the intrinsic assumptions made in most existing IRL techniques.

We investigate IRL in partially observable Markov decision processes (POMDPs), a widely used model for decision-making under imperfect information. 
Partial observability brings two key challenges in IRL.
The first challenge is due to the so-called \emph{information asymmetry} between the expert and the learner. The expert typically has either full or partial information about the environment, while the learner has only a partial view of the state and the expert's demonstrations.
Even in the hypothetical case in which the underlying reward function is known to the learner, its optimal policy under limited information may not yield the same behavior as an expert with full information due to such information asymmetry.

The second challenge is due to the computational complexity of policy synthesis in POMDPs.
Indeed, many standard IRL techniques rely on a subroutine that solves the so-called \emph{forward problem}, i.e., computing an optimal policy for a given reward.
Solving the forward problem for POMDPs is significantly more challenging than MDPs, both theoretically and practically~\cite{vlassis2012computational}. 
Optimal policies for POMDPs may require infinite memory of observations~\cite{MadaniHC99}, whereas memoryless policies are enough for MDPs.

An additional limitation in existing IRL techniques is due to the limited expressivity and often impracticability of state-based reward functions in representing complex tasks~\cite{littman2017environment}.
For example, it will be tremendously difficult to define a merely state-based reward function to describe requirements such as ``do not steer off the road while reaching the target location and coming back to home'' or ``monitor multiple locations with a certain order''.
However, such requirements can be concisely and precisely specified in temporal logic~\cite{BK08,Pnueli}.
Therefore, recent work has demonstrated the utility of incorporating temporal logic specifications into IRL in MDPs~\cite{memarian2020active,wen2017learning}.

In this work, we address these challenges and limitations in state-of-the-art IRL techniques by investigating the following problem.
\begin{mdframed}[backgroundcolor=gray!30,  linecolor=black!60]
\textbf{Task-Guided IRL in POMDPs:} Given a POMDP, a set of expert demonstrations, and, if available, a \emph{task specification} expressed in temporal logic, learn a policy along with the underlying reward function that maximizes the \emph{causal entropy} of the induced stochastic process, induces a behavior similar to the expert's, and ensures the satisfaction of the specification.
\end{mdframed}

We highlight two parts of the problem statement.
Using \emph{causal entropy} as an optimization criterion instead of traditional entropy results in a least-committal policy that induces a behavior obtaining the same accumulated reward as the expert's demonstrations while making no additional assumptions about the demonstrations.
\emph{Task specifications} given as task requirements guide the learning process by describing the feasible behaviors and allow the learner to learn performant policies with respect to the task requirements.
Such specifications can be interpreted as side information available to the learner a priori in addition to the demonstrations aimed at partially alleviating the information asymmetry between the expert and the learner.

\myett{Specifically, we tackle the IRL on POMDPs problem by a reformulation into a maximum causal entropy (MCE) problem. Then, we develop a new solver for the MCE problem that improves computational tractability over existing approaches. The developed solver can enforce prior task knowledge expressed as temporal logic specifications, which guides the learning, improves the data efficiency, and partially alleviates the information asymmetry problem.}

Most existing work on IRL relies on \emph{entropy} as a measure of the likelihood of the demonstrations, yet,  when applied to stochastic MDPs, has to deal with nonconvex optimization problems~\cite{ziebart2008maximum,ziebart2010modeling}.
On the other hand, IRL techniques that adopt \emph{causal entropy} as the measure of likelihood enjoy formulations based on convex optimization~\cite{zhou2017infinite,ziebart2010modeling,ziebart2013principle}.
We show similar algorithmic benefits in maximum-causal-entropy IRL carry over from MDPs to POMDPs.

A major difference between MDPs and POMDPs in maximum-causal-entropy IRL is, though, due to the intrinsic nonconvexity of policy synthesis in POMDPs, which yields a formulation of the task-guided IRL problem as a nonconvex optimization problem.
It is known that this nonconvexity severely limits the scalability for synthesis in POMDPs~\cite{vlassis2012computational}.
We develop an iterative algorithm that solves the resulting nonconvex problem in a scalable manner by adapting sequential convex programming (SCP)~\cite{yuan2015recent,mao2018successive}.
In each iteration, it linearizes the underlying nonconvex problem around the solution from the previous iteration.
The algorithm introduces several extensions to alleviate the errors resulting from the linearization.
One of these extensions is a verification step not present in existing SCP schemes.
We show that the proposed algorithm computes a sound and locally optimal solution to the task-guided problem.

Additionally, we empirically demonstrate that the algorithm scales to POMDPs with tens of thousands of states as opposed to tens of states in most existing work.
In POMDPs, \emph{finite-memory} policies that are functions of the history of the observations outperform memoryless policies~\cite{yu2008near}.
Besides, computing a finite-memory policy for a POMDP is equivalent to computing a memoryless policy on a larger product POMDP~\cite{junges2018finite}.
Thus, we leverage the scalability of our algorithm to compute more performant policies that incorporate memory using finite-state controllers~\cite{meuleau1999solving,amato2010optimizing}.
On the other hand, the existing IRL techniques on POMDPs aforementioned cannot effectively utilize memory, as they do not scale to large POMDPs.

We demonstrate the applicability of the approach through several examples, including a simulated wheeled ground robot operating in a high-fidelity, continuous, 3-D Unity simulation.
We show that, without task specifications, the developed algorithm can compute more performant policies than a straight adaptation of the original GAIL~\cite{ho2016generative} to POMDPs.
Then, we demonstrate that by incorporating task specifications into the IRL procedure, the learned reward function and policy accurately describe the behavior of the expert while outperforming the policy obtained without the task specifications.
We observe that with more limited data, the performance gap becomes more prominent between the learned policies with and without using task specifications.
Most importantly, we empirically demonstrate the scalability of our approach for solving the \emph{forward problem} through extensive comparisons with several state-of-the-art POMDP solvers and show that on larger POMDPs, the algorithm can compute more performant policies in significantly less time.
\section{Preliminaries}
The following section provides a review of prerequisite understanding for POMDPs, their accompanying policies and how a POMDP's belief over states is updated using Bayesian techniques.

\paragraph{\textbf{Notation}} We denote the set of nonnegative real numbers by $\Real_+$, the set of all probability distributions over a finite or countably infinite set $\mathcal{X}$ by $\Distr(\mathcal{X})$, the set of all (infinite or empty) sequences $x_0,x_1,\hdots,x_\infty$ with $x_i \in \mathcal{X}$ by $(\mathcal{X})^*$ for some set $\mathcal{X}$, and the expectation of a function $g$ of jointly distributed random variables $X$ and $Y$ by $\mathbb{E}_{X,Y}[g(X,Y)]$.%

\subsection{Partially Observable Markov Decision Process}\label{subsec:pomdp}
A partially observable Markov decision process (POMDP) is a framework for modeling sequential interaction between an agent and a partially observable environment, where the agent cannot perceive its underlying state but must infer it based on the given noisy observation. 

\paragraph{\textbf{POMDPs}} We define a POMDP by a tuple $\Pomdp = (\States, \Actions, \Transition, \ObservationSet, \Observations, \Reward, \Initdist, \discount)$, where $\States$, $\Actions$, and $\myett{\ObservationSet}$ are finite sets of states, actions, and observations, respectively. The function $\Initdist : \States \mapsto \Real_+$ provides the initial distribution of the agent's state and $\discount \in [0,1)$ is a discount factor over a possibly infinite planning horizon. At each decision time, an agent selects an action $\action \in \Actions$ and the transition function $\Transition : \States \times \Actions \mapsto \Distr(\States)$ defines the probability $\Transition(\state' | \state,\action)$ of reaching state $\state' \in \States$ given the current state $\state \in \States$ and action $\action$. After the state transition, the agent receives an observation $\observation' \in \myett{\ObservationSet}$ according to the function $\Observations : \States \mapsto \Distr(\ObservationSet)$, which defines the probability $\Observations(\observation'|\state')$ of perceiving $\observation'$ at state $\state'$. The agent also receives a reward function $\Reward(\state,\action)$ from the function $\Reward: \States \times \Actions \mapsto \Real$ encoding the task specification.
In the following, without loss of generality, we consider infinite-horizon problems.

\paragraph{\textbf{Policies}}
An observation-based policy $\policy : (\ObservationSet \times \Actions)^* \times \ObservationSet \mapsto \Distr(\Actions)$ for a POMDP $\Pomdp$ maps a sequence of observations and actions to a distribution over actions.
A $\mathrm{M}$-\emph{finite-state controller} ($\mathrm{M}$-FSC) is a tuple $\mathcal{C} = (\mathcal{Q}, q_I, \eta, \delta)$, where $Q = \{q_1,q_2,\hdots,q_M\}$ is a finite set of memory states, $q_I$ is the initial memory state, $\eta : \mathcal{Q} \times \ObservationSet \mapsto \Distr(\Actions)$ is a decision function, and $\delta : \mathcal{Q} \times \ObservationSet \times \Actions \mapsto \Distr(\mathcal{Q})$ is a memory transition function. The \emph{action mapping} $\mapping(n,\observation)$ takes a FSC memory state $n$ and an observation $\observation \in \ObservationSet$, and returns a distribution over the POMDP actions.
The \emph{memory update} $\memoryupt(n,\observation,\action)$ returns a distribution over memory states and is a function of the action $\action$ selected by $\mapping$. An FSC induces an observation-based policy by following a joint execution of these two functions upon a trace of the POMDP. An FSC is \emph{memoryless} if there is a single memory state. \emph{Memoryless FSCs, denoted by $\policy\colon \ObservationSet \rightarrow\Distr(\Actions)$, are observation-based policies, where $\policy(\action|\observation) = \policy_{\observation,\action}$ is the probability of taking the action $\action$ given solely observation $\observation$.}

\begin{remark}[\textsc{Reduction to Memoryless Policies}]\label{rem:mempolicy}
In the remainder of the paper, for ease of notation, we synthesize optimal $\mathrm{M}$-FSCs for POMDPs (so-called forward problem) by computing memoryless policies $\policy$ on theoretically-justified larger POMDPs obtained from the so-called product of the memory update $\memoryupt$ and the original POMDPs. Indeed, the authors of~\cite{junges2018finite} provide product POMDPs, whose sizes grow polynomially only with the size of the domain of $\memoryupt$.
\end{remark}

\paragraph{\textbf{Belief Update}} Given \myett{a history} on the POMDP $\Pomdp$ as the perceived observation and executed action sequence $\tau = \{(\observation_0,\action_0),(\observation_1,\action_1),\hdots,(\observation_{T},\action_{T} )\}$, where $\observation_i \in \ObservationSet$, $\action_i \in \Actions$, $i \in \{0,\hdots,T\}$ and $T$ is the length of the trajectory, the belief state specifies the probability of being in each state of the POMDP given an initial belief $b_0 = \Initdist$. Such a belief state can be updated at each time step using the following Bayes rule
\begin{align}
    b_{t+1}(\state') = \frac{\Observations(\observation_{t} | \state') \sum_{\state\in \States} \Transition(\state' | \state,\action_t) b_t(\state)}{\sum_{\state'' \in \States} \Observations(\observation_t | \state'') \sum_{\state \in \States} \Transition(\state'' | \state,\action_t)b_t(\state)}.\label{eq:bayesian-update}
\end{align}

\subsection{Causal Entropy in POMDPs.}\label{subsec:causalent}

For a POMDP $\Pomdp$, a policy $\policy$ induces the stochastic processes $S^\policy_{0:\infty} := (S^\policy_0,\hdots,S^\policy_\infty) $, $A^\policy_{0:\infty} := (A^\policy_0,\hdots,A^\policy_\infty)$, and $Z^\policy_{0:\infty} := (Z^\policy_0,\hdots,Z^\policy_\infty)$. 
At each time index $t$, the random variables $S^\policy_t$, $A^\policy_t$, and $Z^\policy_t$ take values $\state_t \in \States$, $\action_t \in \Actions$, and $\observation_t \in \ObservationSet$, respectively. The probability $P(A_{0:T} || S_{0:T})$ of $A_{0:T}$ \emph{causally-conditioned} on $S_{0:T}$, given by~\cite{ziebart2010modeling,massey1990causality,kramer1998directed} $P(A_{0:T} || S_{0:T}) := \prod_{t=0}^T P(A_t | S_{0:t}, A_{0:t-1})$, defines a correlation between the stochastic processes, where each variable $A_t$ is conditionally influenced by only the earlier predicted variables $S_{0:t}, A_{0:t-1}$, and not the future variables $S_{t+1:T}$. Let $H(A|S) \triangleq \mathbb{E}_{A,S}[- \log P(A | S)]$ be the \emph{conditional entropy} of a random variable $A$ given a random variable $S$. In the finite-horizon setting, the causal entropy $H_\policy$ induced by a given policy $\policy$ is defined as $H_\policy  := \mathbb{E}_{A^\policy_{0:T},S^\policy_{0:T}}[- \log \mathbb{P}(A^\policy_{0:T} || S^\policy_{0:T})] = \sum_{t=0}^T H(A^\policy_t | S^\policy_{0:t}, A^\policy_{0:t-1})$. Then, the \emph{causal entropy} in the infinite-horizon setting, namely the \emph{discounted causal entropy}~\cite{zhou2017infinite,hansen2006robust}, is defined as
\begin{align}
    H_\policy^\discount := \sum\nolimits_{t=0}^\infty \discount^t H(A^\policy_t | S^\policy_{0:t}, A^\policy_{0:t-1}) = \sum\nolimits_{t=0}^\infty \discount^t \mathbb{E}_{A^\policy_t,S^\policy_t}[- \log \mathbb{P}(A^\policy_t | S^\policy_t)], \label{eq:discounted-causal-entropy}
\end{align}%
where the second equality is due to the Markov property.

\begin{remark}
The \emph{entropy} of POMDPs (or MDPs) involves the \emph{future policy decisions}~\cite{ziebart2008maximum}, i.e., $S^\policy_{t+1:T}$, at a time index $t$, as opposed to the causal entropy in POMDPs (or MDPs). Thus, the authors of \cite{ziebart2008maximum} show that the problem of computing a policy that maximizes the entropy is nonconvex, even in MDPs.
Inverse reinforcement learning techniques that maximize the entropy of the policy rely on approximations or assume that the transition function of the MDP is deterministic.
On the other hand, computing a policy that maximizes the causal entropy can be formulated as a convex optimization problem in MDPs~\cite{ziebart2010modeling,zhou2017infinite}.
\end{remark}

\subsection{LTL Specifications.} 

We use general linear temporal logic (LTL) to express complex task specifications on the POMDP $\Pomdp$. Given a set $\mathrm{AP}$ of atomic propositions, i.e., Boolean variables with truth values for a given state $s$ or observation $z$, LTL formulae are constructed inductively as following: $$\reachPropSymbol := \mathrm{true} \;|\; a \;|\; \lnot \reachPropSymbol \;|\; \reachPropSymbol_1 \wedge \reachPropSymbol_2 \;|\; \textbf{X} \reachPropSymbol \;|\; \reachPropSymbol_1 \textbf{U} \reachPropSymbol_2,$$ where $a \in \mathrm{AP}$, $\reachPropSymbol$, $\reachPropSymbol_1$, and $\reachPropSymbol_2$ are LTL formulae, $\lnot$ and $\wedge$ are the logic negation and conjunction, and $\textbf{X}$ and $\textbf{U}$ are the \emph{next} and \emph{until} temporal operators. Besides, temporal operators such as \emph{always} ($\textbf{G}$) and \emph{eventually} ($\textbf{F}$) are derived as $\textbf{F}\reachPropSymbol := \mathrm{true} \textbf{U} \reachPropSymbol$ and $\textbf{G} \reachPropSymbol := \lnot \textbf{F} \lnot\reachPropSymbol$. We denote by \emph{$\mathrm{Pr}_{\Pomdp}^\policy(\varphi)$ the probability of satisfying the LTL formula $\varphi$ when following the policy $\sigma$ on the POMDP $\Pomdp$.} A detailed description of the syntax and semantics of LTL is beyond the scope of this paper and can be found in~\cite{Pnueli,BK08}.
\section{Problem Formulation}

In this section, we formulate the problem of task-guided inverse reinforcement learning (IRL) in POMDPs. Given a POMDP $\Pomdp$ with an \emph{unknown} reward function $\Reward$, we seek to learn a reward function $\Reward$ along with an underlying policy $\policy$ that induces a behavior similar to the expert demonstrations. 

We define an expert trajectory on the POMDP $\Pomdp$ as the perceived observation and executed action sequence $\tau = \{(\observation_0,\action_0),(\observation_1,\action_1),\hdots,(\observation_{T},\action_{T} )\}$, where $\observation_i \in \ObservationSet$ and $\action_i \in \Actions$ for all $i \in \{0,\hdots,T\}$, and $T$ denotes the length of the trajectory. Similarly to \cite{choi2011inverse}, we assume given or we can construct from $\tau$ (via Bayesian belief updates~\eqref{eq:bayesian-update}) the belief trajectory $b^{\tau} = \{b_{0}:=\mu_0,\hdots,b_{T}\}$, where $b_{i}(\state)$ is the estimated  probability of being at state $\state$ at time index $i$. In the following, we assume that we are given a set of belief trajectories $\mathcal{D} = \{b^{\tau_1},\hdots, b^{\tau_N}\}$ from trajectories $\tau_1,\hdots,\tau_N$, where $N$ denotes the total number of underlying trajectories.

We parameterize the unknown reward function $\Reward$ by a differentiable function (with respect to the parameter) $\Reward^\Rewardvec : \States \times \Actions \mapsto \Real^d$, where $\Rewardvec \in \mathbb{R}^F$ is a parameter that defines uniquely the reward function.  Such an encoding includes traditional \myett{representations} of the reward such as $\Reward^\Rewardvec(\state,\action) = g_\Rewardvec(\phi(\state, \action))$, where $\phi : \States \times \Actions \mapsto \Real^d$ is a known vector of basis functions with components referred to as \emph{feature functions}, $d$ is the number of features, and $g_\Rewardvec$ can be any function approximator such as  neural networks. For example, in the traditional linear encoding, we have $g_\Rewardvec(z) = \Rewardvec^{\mathrm{T}} z$.

Specifically, we seek for a parameter $\Rewardvec$ defining $\Reward^\Rewardvec$ and a policy $\policy$ such that its discounted return expectation $R_\policy^\Rewardvec$ matches an empirical discounted return expectation $\bar{R}^\Rewardvec$ of the expert demonstration $\mathcal{D}$. That is, we have that $R_\policy^\Rewardvec = \bar{R}^\Rewardvec$, where
\begin{align*}
    R_\policy^\Rewardvec := \sum_{t=0}^\infty \discount^t \mathbb{E}_{S^\policy_t,A^\policy_t}[\Reward^\Rewardvec(S^\policy_t,A^\policy_t)| \policy] \; \text{and} \; \bar{R}^\Rewardvec = \frac{1}{N} \sum_{b^{\tau} \in \mathcal{D}} \sum_{b_i \in b^{\tau} } \discount^i \sum_{\state \in \States}b_i(s) \Reward^\Rewardvec(s,\action_i).
\end{align*}
In the case of linear encoding of the reward, the above condition is called feature matching expectation, and it can be simplified by replacing $\Reward^\Rewardvec$ with the feature function $\phi$.

Nevertheless, the problem is ill-posed and there may be infinitely many reward functions and policies that can satisfy the above matching condition.
To resolve the ambiguities, we seek for a policy $\policy$ that also maximizes the discounted causal entropy $H_\policy^\discount$.  We now define the problem of interest.
\begin{problem}\label{pb:irl-pomdp}
    Given a reward-free POMDP $\Pomdp$, a demonstration set $\mathcal{D}$, and a feature $\phi$, compute a policy $\policy$ and weight $\Rewardvec$ such that (a) The matching condition holds; (b) The causal entropy $H_\policy^\discount$ given by~\eqref{eq:discounted-causal-entropy} is maximized by $\policy$.
\end{problem}%

Furthermore, we seek to incorporate, if available, a priori high-level side information on the task demonstrated by the expert in the design of the reward and policy.%
\begin{problem}\label{pbirl-pomdp-side-info}
    Given a linear temporal logic formula $\varphi$, compute a policy $\policy$ and weight $\Rewardvec$ such that the constraints (a) and (b) in Problem~\ref{pb:irl-pomdp} are satisfied, and $\mathrm{Pr}_{\Pomdp}^\policy(\varphi) \geq \lambda$ for a given parameter $\lambda \geq 0$.
\end{problem}

Although the parameter $\lambda$ that specifies the threshold for satisfaction of $\reachPropSymbol$ is assumed to be given, the approach can easily be adapted to compute the optimal $\lambda$.
\section{Nonconvex Formulation for IRL in POMDPs}

In this section, we formulate Problem~\ref{pb:irl-pomdp} and Problem ~\ref{pbirl-pomdp-side-info} as finding saddle points of a nonconvex functions. 
Then, we propose an algorithm based on solving a nonconvex optimization problem to compute such saddle points. We \myett{emphasize} (see Remark~\ref{rem:mempolicy}) that we compute $\mathrm{M}$-FSC for POMDPs by computing memoryless policies $\policy$ on larger product POMDPs. Indeed, in the remainder of the paper, we reason directly on the product POMDP, which is the product of a POMDP and an FSC, and it yields a POMDP with state memory pairs~\cite{junges2018finite}.

\paragraph{\textbf{Substituting Visitation Counts}} We eliminate the (infinite) time dependency in $H_\policy^{\discount}$ and the matching condition by a substitution of variables involving the policy-induced \emph{discounted state visitation count} $\StateOccup^\discount_{\policy} : \States \mapsto \Real_+$ and \emph{state-action visitation count} $\StateActionOccup^\discount_{\policy} : \States \times \Actions \mapsto \Real_+$. 
For a policy $\policy$, state $\state$, and action $\action$, the discounted state and state-action visitation counts are defined by 
\begin{align*}
    \StateOccup^\discount_\policy (\state) := \mathbb{E}_{S_t} [\sum_{t=1}^\infty \discount^t \mathbbm{1}_{\{S_t = \state\}} | \policy] \; \text{and} \; \StateActionOccup^\discount_\policy (\state,\action) := \mathbb{E}_{A_t, S_t} [\sum_{t=1}^\infty \discount^t \mathbbm{1}_{\{S_t = \state, A_t = \action\}} | \policy],
\end{align*}
where $\mathbbm{1}_{\{\cdot\}}$ is the indicator function. From these definitions, it is straightforward to deduce that $\StateActionOccup^\discount_\policy(\state,\action) = \pi_{\state,\action}\StateOccup^\discount_\policy(\state)$, where $\pi_{\state,\action} = \mathbb{P}[A_t=a | S_t = s] $. It is also straightforward to check that for all $\state \in \States$ and $\action \in \Actions$,  $\StateOccup^\discount_{\policy} (\state) \geq 0$, $\StateActionOccup^\discount_{\policy} (\state,\action) \geq 0$, and $\StateOccup^\discount_{\policy}(\state) = \sum_{\action \in \Actions} \StateActionOccup^\discount_{\policy}(\state,\action).$

We first provide a concave expression for the discounted causal entropy $H_\policy^\discount$ as a function of the visitation counts $\StateOccup^\discount_\policy$ and $\StateActionOccup^\discount_\policy$:%
\begin{align}
    H_\policy^\discount 
    &:= \sum\nolimits_{t=0}^\infty \discount^t \mathbb{E}_{S^\policy_t,A^\policy_t}[-\log(\pi_{\state_t,\action_t})] \nonumber \\
    &= \sum\nolimits_{t=0}^\infty \sum\nolimits_{(\state,\action) \in \States \times \Actions} -(\log \pi_{\state,\action}) \pi_{\state,\action} \discount^t \mathbb{P}[S^\policy_t=s] \nonumber\\
&= \sum\nolimits_{(\state,\action)\in \States \times \Actions}  -(\log \pi_{\state,\action}) \pi_{\state,\action} \StateOccup^\discount_\policy (\state) \nonumber\\
&= \sum\nolimits_{(\state,\action)\in \States \times \Actions} -\log \frac{\StateActionOccup^\discount_\policy(\state,\action)}{\StateOccup^\discount_\policy(\state)} \StateActionOccup^\discount_\policy(\state,\action), \label{eq:causal-entropy}
\end{align}%
where the first equality is due to the definition of the discounted causal entropy $H_\policy^\discount$, the second equality \myett{is} obtained by expanding the expectation. 
The third and fourth equalities follow by the definition of the state visitation count $\StateOccup^\discount_{\policy}$, and the state-action visitation count $\StateActionOccup^\discount_{\policy}$. We prove in the appendix that the above expression is indeed concave in the visitation counts.
Next, we obtain a \emph{linear} expression in $\StateActionOccup^\discount_\policy$ for the discounted return expectation $R^\Rewardvec_\policy$ as:%
\begin{align}
    R^\Rewardvec_\policy \: &= \: \sum_{t=0}^\infty \sum_{(\state,\action) \in \States \times \Actions} \Reward^\Rewardvec(\state,\action) \discount^t \mathbb{P}[S^\policy_t = \state, A^\policy_t = \action] \nonumber \\
    &= \: \sum_{(\state,\action) \in \States \times \Actions} \Reward^\Rewardvec(\state,\action) \StateActionOccup^\discount_\policy(\state,\action), \label{eq:feature-matching}
\end{align} where the second equality is obtained by the definition of the visitation count $\StateActionOccup^\discount_{\policy}$. The following \emph{nonconvex} constraint in $\StateOccup^\discount_\policy(\state)$ and $\policy_{\observation,\action}$ ensures observation-based policies:%
\begin{align}
 \StateActionOccup^\discount_\policy (\state,\action)  =  \StateOccup^\discount_\policy (\state) \sum\nolimits_{\observation \in \ObservationSet}\Observations(\observation|\state)\policy_{\observation,\action}. \label{eq:policy-constraint}
\end{align}%
Finally, the variables for the discounted visitation counts must satisfy the so-called \emph{Bellman flow constraint} \myett{\cite{zhou2017infinite}} to ensure that the policy is well-defined. For each state $s \in \States$,%
\begin{align}
    \StateOccup^\discount_{\policy} (\state)  = \Initdist(\state) + \discount \sum_{\state' \in \States} \sum_{\action \in \Actions} \Transition(\state | \state',\action) \StateActionOccup^\discount_{\policy} (\state',\action). \label{eq:bellman-constraint}
\end{align}%

\algdef{SE}[DOWHILE]{Do}{doWhile}{\algorithmicdo}[1]{\algorithmicwhile\ #1}%
\begin{algorithm}[t]
    \caption{Compute the weight vector $\Rewardvec$ and policy $\policy$ solution of the Lagrangian relaxation of the IRL problem. }\label{algo:gradweight}
    \begin{algorithmic}[1]    
        \Require{Feature expectation $\bar{R}^\phi$ from $\mathcal{D}$, initial weight $\Rewardvec^0$, step size $\eta : \mathbb{N} \mapsto \mathbb{R}^+$, and (if available) a priori side information $\varphi$ and $\lambda \in [0,1]$ imposing $\mathrm{Pr}_{\Pomdp}^\policy(\varphi) \geq \lambda$ .}
        \State $\policy^0 \gets$ uniform policy \Comment{Initialize uniform policy}
        \For{$k = 0,1,\hdots, $} \Comment{Compute $\Rewardvec$ via gradient descent}
            \State $\sigma^{k+1} \gets \texttt{SCPForward}(\Rewardvec^{k},\policy^{k}, \varphi, \lambda)$ \Comment{Solve the \emph{forward problem}~\eqref{eq:cost-entropy-npb}--\eqref{eq:pos-mu-state-npb} with optional $\varphi$ and $\lambda$ }
            \State $\Rewardvec^{k+1} \gets \Rewardvec^{k} - \eta(k) \nabla_\Rewardvec f(\Rewardvec^k; \policy^{k+1})$ \Comment{Gradient step}
        \EndFor
    \State \Return $\policy^k$, $\Rewardvec^{k}$
  \end{algorithmic}
\end{algorithm}

\paragraph{\textbf{Saddle Point Formulation}} Computing a policy $\policy$ that satisfies the return matching constraint $R^\Rewardvec_\policy = \bar{R}^\Rewardvec$ might be infeasible due to $\bar{R}^\Rewardvec$ being an empirical estimate from the finite set of demonstrations $\mathcal{D}$. 
Additionally, the feature matching constraint might also be infeasible due to the information asymmetry between the expert and the learner, e.g., the expert has full observation.

We build on a saddle point computation problem to incorporate the return matching constraints into the objective of the forward problem, \myett{similar to} other IRL algorithms in the literature. 
Specifically, the desired weight vector $\Rewardvec$ and policy $\policy$ of Problem~\ref{pb:irl-pomdp} and Problem~\ref{pbirl-pomdp-side-info} are the solutions of $\mathrm{min}_\Rewardvec \: f(\Rewardvec) := \mathrm{max}_\policy \: H_\policy^\discount + (R_\policy^\Rewardvec - \bar{R}^\Rewardvec)$. \myett{The function $f$ corresponds to the inner optimization problem when the reward parameter is fixed. That is, $f(\theta)$ computes a policy $\sigma$ that maximizes the sum $H_\policy^\discount + R_\policy^\Rewardvec$ of the causal entropy and the current estimate of the reward function. In other words, $f(\theta)$ returns the solution to the forward problem, i.e., finding optimal policy on the POMDP when the entropy term is removed.}

Algorithm~\ref{algo:gradweight} updates the reward weights by using gradient descent. Initially, the policy $\policy^0$ is a random uniform variable and the weight $\Rewardvec^0$ is a nonzero vector. At iteration $k\geq 0$, the policy $\policy^{k+1} = \argmax_\policy H_\policy^\discount +  (R_\policy^{\Rewardvec^{k} } - \bar{R}^{\Rewardvec^{k} })$ is the optimal policy on the POMDP under the current reward estimate $\Reward^{\Rewardvec^{k} }$ given by $\Rewardvec^{k} $. That is, $\policy^{k+1}$ is the solution to the \emph{forward problem}. Then, to update the weight $\Rewardvec$, Algorithm~\ref{algo:gradweight} computes the gradient $\nabla_\Rewardvec f$ with respect to $\Rewardvec$ as follows:
\begin{align}
    \nabla_\Rewardvec f (\Rewardvec; \policy) = \sum_{\state,\action \in \States \times \Actions} \StateActionOccup^\discount_{\policy}(\state,\action) \nabla_\Rewardvec\Reward^\Rewardvec(\state,\action) - \frac{1}{N} \sum_{b^{\tau} \in \mathcal{D}} \sum_{b_i \in b^{\tau} } \discount^i \sum_{\state \in \States}b_i(s) \nabla_\Rewardvec\Reward^\Rewardvec(s,\action_i). \nonumber
\end{align}

We develop the algorithm $\texttt{SCPForward}$, presented in next section, to solve the forward problem, i.e., computing $\policy^{k+1}$ given $\Rewardvec^{k}$, in an efficient and scalable manner while incorporating high-level task specifications to guide the learning.

\paragraph{\textbf{Nonconvex Formulation of the Forward Problem}} Given a weight vector $\Rewardvec^k$, we take advantage of the obtained substitution by the expected visitation counts to formulate the \emph{forward problem} associated to Problem~\ref{pb:irl-pomdp} as the nonconvex optimization problem:
\begin{align}
    &\displaystyle \underset{\StateOccup^\discount_\policy,\StateActionOccup^\discount_\policy,\policy}{\mathrm{maximize}}   \: \sum_{(\state,\action)\in \States \times \Actions} -\log \frac{\StateActionOccup^\discount_\policy(\state,\action)}{\StateOccup^\discount_\policy(\state)} \StateActionOccup^\discount_\policy(\state,\action) + \sum_{(\state,\action) \in \States \times \Actions} \Reward^{\Rewardvec^{k}}(\state,\action) \StateActionOccup^\discount_\policy(\state,\action) \label{eq:cost-entropy-npb}\\ 
    &\displaystyle \mathrm{subject \ to} \quad\eqref{eq:policy-constraint}-\eqref{eq:bellman-constraint},\nonumber\\
    &\displaystyle\forall (\state,\action) \in \States \times \Actions, \;\; \StateOccup^\discount_{\policy} (\state) \geq 0,  \;\; \StateActionOccup^\discount_\policy (\state,\action) \geq 0, \label{eq:pos-and-mu-state-npb}\\
    & \displaystyle\forall (\state,\action) \in \States \times \Actions, \;\; \StateOccup^\discount_{\policy}(\state) = \sum\nolimits_{\action \in \Actions} \StateActionOccup^\discount_{\policy}(\state,\action), \label{eq:pos-mu-state-npb}
\end{align}%
where the source of nonconvexity is from~\eqref{eq:policy-constraint}, and we remove the constant $-\bar{R}^{\Rewardvec^k}$ from the cost function of the above optimization problem.
\section{Sequential Linear Programming Formulation}
We develop an algorithm, $\texttt{SCPForward}$, adapting a sequential convex programming (SCP) scheme to efficiently solve the nonconvex \emph{forward problem}~\eqref{eq:cost-entropy-npb}--\eqref{eq:pos-mu-state-npb}. Indeed, $\texttt{SCPForward}$ involves a \emph{verification step} to compute sound policies and visitation counts, which is not present in the existing SCP schemes. Additionally, we describe in the next section how to take advantage of high-level task specification (Problem~\ref{pbirl-pomdp-side-info}) through slight modifications of the obtained optimization problem solved by $\texttt{SCPForward}$.

\subsection{Linearizing Nonconvex Optimization Problem}
$\texttt{SCPForward}$ iteratively linearizes the nonconvex constraints in~\eqref{eq:policy-constraint} around a previous solution.
However, the linearization may result in an infeasible or unbounded linear subproblem~\cite{mao2018successive}. 
We first add \emph{slack variables} to the linearized constraints to ensure feasibility.
The linearized problem may not accurately approximate the nonconvex problem if the solutions to this problem deviate significantly from the previous solution.
Thus, we utilize trust region constraints~\cite{mao2018successive} to ensure that the linearization is accurate to the nonconvex problem. 
At each iteration, we introduce a \emph{verification step} to ensure that the computed policy and visitation counts are not just approximations but actually satisfy the nonconvex policy constraint~\eqref{eq:policy-constraint}, improves the realized cost function over past iterations, and satisfy the temporal logic specifications, if available.

\paragraph{\textbf{Linearizing Nonconvex Constraints and Adding Slack Variables}} We linearize the nonconvex constraint~\eqref{eq:policy-constraint}, which is quadratic in $\StateOccup^\discount_\policy(\state)$ and $\policy_{\observation,\action}$, around the previously computed solution denoted by $\hat{\policy}$, $\StateOccup^\discount_{\hat{\policy}}$, and $\StateActionOccup^\discount_{\hat{\policy}}$. 
However, the linearized constraints may be infeasible.
We alleviate this drawback by adding \emph{slack variables} $\slack_{\state,\action}\in \Real$ for $(\state,\action) \in \States \times \Actions$, which results in the affine constraint:%
\begin{align}
 \StateActionOccup^\discount_\policy (\state,\action) + \slack_{\state,\action} =& \;\StateOccup^\discount_{\hat{\policy}} (\state) \sum\nolimits_{\observation \in \ObservationSet}\Observations(\observation|\state){\policy}_{\observation,\action} \;+ \label{eq:convex-policy-constraint-slack} \\
    &\big(\StateOccup^\discount_\policy (\state)-\StateOccup^\discount_{\hat{\policy}} (\state)\big) \sum\nolimits_{\observation \in \ObservationSet}\Observations(\observation|\state)\hat{\policy}_{\observation,\action}. \nonumber
\end{align}%

\paragraph{\textbf{Trust Region Constraints}} The linearization may be inaccurate if the solution deviates significantly from the previous solution.
We add following \emph{trust region} constraints to alleviate this drawback:%
\begin{align}
     \forall (\observation,\action) \in \ObservationSet \times \Actions,& \quad \hat{\policy}_{\observation,\action}/\trustregion \leq {\policy}_{\observation,\action} \leq  \hat{\policy}_{\observation,\action} \trustregion, \label{eq:convex-policy-constraint-trust-region}
\end{align}%
where $\trustregion$ is the size of the trust region to restrict the set of allowed policies in the linearized problem. We augment the cost function in \eqref{eq:cost-entropy-npb} with the term $-\beta \sum_{(\state,\action) \in \States\times\Actions}\slack_{\state,\action}$ to ensure that we minimize the violation of the linearized constraints, where $\beta$ is a large positive constant.

\paragraph{\textbf{Linearized Problem}} Finally, by differentiating $x \mapsto x \log x$ and $y \mapsto x \log (x/y)$, we obtain the coefficients required to linearize the convex causal entropy cost function in~\eqref{eq:cost-entropy-npb}. Thus, we obtain the following linear program (LP):%
\begin{align}
    & \underset{\StateOccup^\discount_\policy,\StateActionOccup^\discount_\policy,\policy}{\mathrm{maximize}}   \: \sum\nolimits_{(\state,\action)\in \States \times \Actions} -\Bigg(\beta\slack_{\state,\action}- \Big(\frac{\StateActionOccup^\discount_{\hat{\policy}}(\state,\action)}{\StateOccup^\discount_{\hat{\policy}}(\state)} \Big)\StateOccup^\discount_\policy(\state) \nonumber \\ 
    & \quad \quad \quad \quad \quad \: + \Big(\log \frac{\StateActionOccup^\discount_{\hat{\policy}}(\state,\action)}{\StateOccup^\discount_{\hat{\policy}}(\state)}+1 \Big)\StateActionOccup^\discount_\policy(\state,\action) \Bigg) + \sum_{(\state,\action) \in \States \times \Actions} \Reward^{\Rewardvec^{k}}(\state,\action) \StateActionOccup^\discount_\policy(\state,\action) \label{eq:cost-linearized-pb}\\
    & \mathrm{subject \ to} \quad  \eqref{eq:bellman-constraint},\eqref{eq:pos-and-mu-state-npb}-\eqref{eq:convex-policy-constraint-trust-region}.\nonumber
\end{align}

\paragraph{\textbf{Verification Step}} After each iteration, the linearization might be inaccurate, i.e, the resulting policy $\Tilde{\policy}$ and \emph{potentially inaccurate} visitation counts $\Tilde{\StateActionOccup}^\discount_{\Tilde{\policy}}, \Tilde{\StateOccup}^\discount_{\Tilde{\policy}}$ might not be feasible to the nonconvex policy constraint~\eqref{eq:policy-constraint}.
As a consequence of the potential infeasibility, the currently attained (linearized) optimal cost might significantly differ from the \emph{realized cost} by the feasible visiation counts for the $\Tilde{\policy}$.   
Additionally, existing SCP schemes linearizes the nonconvex problem around the previously inaccurate solutions for $\Tilde{\StateActionOccup}^\discount_{\Tilde{\policy}}$, and $\Tilde{\StateOccup}^\discount_{\Tilde{\policy}}$, further propagating the inaccuracy. 
The proposed \emph{verification step} solves these issues. 
Given the computed policy $\Tilde{\sigma}$, \texttt{SCPForward} computes the \emph{unique and sound} solution for the visitation count $\StateOccup^\discount_{\Tilde{\policy}}$ by solving the corresponding \emph{Bellman flow} constraints:
\begin{align}
    \StateOccup^\discount_{\Tilde{\policy}} (\state)  =& \Initdist(\state) + \discount \sum_{\state' \in \States} \sum_{\action \in \Actions} \Transition(\state | \state',\action) \StateOccup^\discount_{\Tilde{\policy}} (\state') \sum_{\observation \in \ObservationSet}\Observations(\observation|\state)\Tilde{\policy}_{\observation,\action}, \label{eq:verif-bellman} 
\end{align}%
for all $\state \in \States$, and where $\StateOccup^\discount_{\Tilde{\policy}} \geq 0$ is the only variable of the linear program. Then, \texttt{SCPForward} computes $\StateActionOccup^\discount_{\Tilde{\policy}}(\state,\action) = \StateOccup^\discount_{\Tilde{\policy}} (\state') \sum_{\observation \in \ObservationSet}\Observations(\observation|\state)\Tilde{\policy}_{\observation,\action}$ and the \emph{\myett{realized cost}} at the current iteration is defined by%
\begin{align}
    \mathrm{C}(\Tilde{\sigma}, \Rewardvec^k) = & \quad \sum_{(\state,\action)\in \States \times \Actions} -\log \frac{\StateActionOccup^\discount_{\Tilde{\policy}}(\state,\action)}{\StateOccup^\discount_{\Tilde{\policy}}} \StateActionOccup^\discount_{\Tilde{\policy}}(\state,\action) +  \sum_{(\state,\action) \in \States \times \Actions} \Reward^{\Rewardvec^{k}}(\state,\action) \StateActionOccup^\discount_{\Tilde{\policy}}(\state,\action),\label{eq:realized-cost}
\end{align}%
where we assume $0 \log 0 = 0$. Finally, if the realized cost $\mathrm{C}(\Tilde{\sigma}, \Rewardvec^k)$ does not improve over the previous cost $\mathrm{C}(\hat{\sigma}, \Rewardvec^k)$, the verification step rejects the obtained policy $\Tilde{\policy}$, contracts the trust region, and $\texttt{SCPForward}$ iterates with the previous solutions $\hat{\policy}$, $\StateOccup^\discount_{\hat{\policy}}$, and $\StateActionOccup^\discount_{\hat{\policy}}$ . 
Otherwise, the linearization is sufficiently accurate, the trust region is expanded, and $\texttt{SCPForward}$ iterates with $\Tilde{\policy}$, $\StateOccup^\discount_{\Tilde{\policy}}$ and $\StateActionOccup^\discount_{\Tilde{\policy}}$.
\emph{By incorporating this verification step, we ensure that $\texttt{SCPForward}$ always linearizes the nonconvex optimization problem around a solution that satisfies the nonconvex constraint~\eqref{eq:policy-constraint}.}%

\subsection{Incorporating High-Level Task Specifications}
Given high-level side information on the agent tasks as the LTL formula $\reachPropSymbol$, we first compute the product of the POMDP and the $\omega$-automaton representing $\reachPropSymbol$ to find the set $\mathcal{T} \subseteq \States$ of states, called target or reach states, satisfying $\varphi$ with probability $1$ by using standard graph-based algorithms as a part of preprocessing step. We refer the reader to ~\cite{BK08} for a detailed introduction on how LTL specifications can be reduced to reachability specifications given by $\mathcal{T}$.

As a consequence, the probability of satisfying $\reachPropSymbol$ is the sum of the probability of reaching the target states $s \in \mathcal{T}$, which are given by the \emph{undiscounted state visitation count} $\StateOccup^\mathrm{sp}_{\policy}$. That is, $\mathrm{Pr}_{\Pomdp}^\policy(\varphi) = \sum_{s \in \mathcal{T}} \StateOccup^\mathrm{sp}_{\policy}(s)$. Unless $\discount = 1$, $\StateOccup^\mathrm{sp}_{\policy} \neq \StateOccup^\discount_{\policy}$. 
Thus, we introduce new variables $\StateOccup^\mathrm{sp}_{\policy}, \StateActionOccup^\mathrm{sp}_{\policy}$, and the adequate constraints in the linearized problem~\eqref{eq:cost-linearized-pb}.

\paragraph{\textbf{Incorporating Undiscounted Visitation Variables to Linearized Problem}} We append new constraints, similar to~\eqref{eq:pos-and-mu-state-npb},~\eqref{eq:pos-mu-state-npb}, and~\eqref{eq:convex-policy-constraint-slack}, into the linearized problem~\eqref{eq:cost-linearized-pb}, where the variables $\StateOccup^\discount_{\policy}, \StateActionOccup^\discount_{\policy}, \slack_{\state,\action}, \StateOccup^\discount_{\hat{\policy}}$, $\StateActionOccup^\discount_{\hat{\policy}}$ are replaced by $\StateOccup^\mathrm{sp}_{\policy}, \StateActionOccup^\mathrm{sp}_{\policy}$, $\slack^{\mathrm{sp}}_{\state,\action}, \StateOccup^{\mathrm{sp}}_{\hat{\policy}}$, $\StateActionOccup^{\mathrm{sp}}_{\hat{\policy}}$, respectively. 
Further, we add the constraint
\begin{align}
    \StateOccup^{\mathrm{sp}}_{\policy} (\state)  = \Initdist(\state) + \sum_{\state' \in \States \setminus \mathcal{T}} \sum_{\action \in \Actions} \Transition(\state | \state',\action) \StateActionOccup^\mathrm{sp}_{\policy} (\state',\action), \label{eq:ltl-bellman}
\end{align}
which is a modification of the \emph{Bellman flow constraints} such that $\StateOccup^\mathrm{sp}_{\policy}(s)$ for all $s \in \mathcal{T}$ only counts transitions from non-target states. 
Finally, we penalize the introduced slack variables for feasibility of the linearization by augmenting the cost function with the term $-\beta \sum_{(\state,\action) \in \States\times\Actions}\slack^{\mathrm{sp}}_{\state,\action}$.

\paragraph{\textbf{Relaxing Specification Constraints}} To incorporate the probability of satisfying the specifications, We add the following constraint to the linearized problem:
\begin{align}
    (\mathrm{spec}) := \sum_{s \in \mathcal{T}} \StateOccup^\mathrm{sp}_{\policy}(s) + \Gamma^\mathrm{sp} \geq \lambda,\label{eq:contr-sat}
\end{align}
where we introduce $\Gamma^\mathrm{sp} \geq 0$ as a slack variable ensuring that the linearized problem is always feasible. 
Further, we augment the cost function with $-\beta^{\mathrm{sp}} \Gamma^\mathrm{sp}$ to penalize violating $\varphi$, where $\beta^{\mathrm{sp}}$ is a positive hyperparameter.%

\paragraph{\textbf{Updating Verification Step}} We modify the previously-introduced realized cost $\mathrm{C}(\Tilde{\sigma}, \Rewardvec^k)$ to penalize when the obtained policy does not satisfy the specification $\varphi$.
This cost also accounts for the linearization inaccuracy of the new policy constraint due to $\policy$, $\StateOccup^\mathrm{sp}_\policy$, and $\StateActionOccup^\mathrm{sp}_\policy$. 
At each iteration, $\texttt{SCPForward}$ computes the accurate $\StateOccup^\mathrm{sp}_{\Tilde{\policy}}$ of current policy $\Tilde{\policy}$ through solving a feasibility LP with constraints given by the \emph{modified Bellman flow constraints}~\eqref{eq:ltl-bellman}. 
Then, it augments $\mathrm{C}^{\mathrm{sp}}_{\Tilde{\policy}} = \min \{0, (\sum_{s \in \mathcal{T}} \StateOccup^\mathrm{sp}_{\Tilde{\policy}}(s) - \lambda) \beta^{\mathrm{sp}}\}$ to the realized cost to take the specification constraints into account.%

\algdef{SE}[DOWHILE]{Do}{doWhile}{\algorithmicdo}[1]{\algorithmicwhile\ #1}%
\begin{algorithm*}[t]
    \caption{\texttt{SCPForward:} Linear programming-based algorithm to solve the \emph{forward problem}~\eqref{eq:cost-entropy-npb}--\eqref{eq:pos-mu-state-npb}, i.e., compute a policy $\policy^{k+1}$ that maximizes the causal entropy, considers the matching constraint, and satisfies the specifications, if available.}\label{algo:scpforward}
    \begin{algorithmic}[1]    
        \Require{Current weight estimate $\Rewardvec^k$, current best policy $\hat{\policy}=\policy^k$, side information $\varphi$ and $\lambda$, trust region $\trustregion > 1$, penalization coefficients $\beta$, $\beta^{\mathrm{sp}} \geq 0$, constant $\trustregion_0$ to expand or contract trust region, and a threshold $\trustregion_{\mathrm{lim}}$ for trust region contraction.}
        \State Find $\StateOccup^\discount_{\hat{\policy}}$ via linear constraint~\eqref{eq:verif-bellman} and $\StateActionOccup^\discount_{\hat{\policy}}=\StateOccup^\discount_{\hat{\policy}} (\state') \sum_{\observation \in \ObservationSet}\Observations(\observation|\state)\hat{\policy}_{\observation,\action}$, given $\hat{\policy}$ \Comment{Realized visitation counts} \label{alg:verif1}
        \State Find $\StateOccup^\mathrm{sp}_{\hat{\policy}}$ via linear constraint~\eqref{eq:ltl-bellman} with $\StateActionOccup^\mathrm{sp}_{\hat{\policy}}=\StateOccup^\mathrm{sp}_{\hat{\policy}} (\state') \sum_{\observation \in \ObservationSet}\Observations(\observation|\state)\hat{\policy}_{\observation,\action}$, given $\hat{\policy}$ \Comment{If $\reachPropSymbol$ is available} \label{alg:verif2}
        \State Compute the realized cost $\mathrm{C}(\hat{\sigma}, \Rewardvec^k) \gets \eqref{eq:realized-cost} + \mathrm{C}^{\mathrm{sp}}_{\hat{\policy}}$, given $\hat{\policy}$ \Comment{Add specifications' violation} \label{alg:verif3}
        \While{$\trustregion > \trustregion_{\mathrm{lim}}$} \Comment{Trust region threshold}
            \State Find optimal $\Tilde{\policy}$ to the augmented LP~\eqref{eq:cost-linearized-pb} via $\hat{\policy}$, $\StateOccup^\discount_{\hat{\policy}}$, $\StateActionOccup^\discount_{\hat{\policy}}$, $\StateOccup^\mathrm{sp}_{\hat{\policy}}$, $\StateActionOccup^\mathrm{sp}_{\hat{\policy}}$ \Comment{We augment the LP with constraints ~\eqref{eq:pos-and-mu-state-npb},~\eqref{eq:pos-mu-state-npb},~\eqref{eq:convex-policy-constraint-slack},~\eqref{eq:ltl-bellman}, and~\eqref{eq:contr-sat} induced by $\StateOccup^\mathrm{sp}_{\policy}, \StateActionOccup^\mathrm{sp}_{\policy}$, and by adding $-\beta \sum_{(\state,\action) \in \States\times\Actions}\slack^{\mathrm{sp}}_{\state,\action} -\beta^{\mathrm{sp}} \Gamma^\mathrm{sp}$ to the cost~\eqref{eq:cost-linearized-pb}.}
            \State Compute the realized $\StateOccup^\discount_{\Tilde{\policy}}$, $\StateActionOccup^\discount_{\Tilde{\policy}}$,$\StateOccup^\mathrm{sp}_{\Tilde{\policy}}$, $\StateActionOccup^\mathrm{sp}_{\Tilde{\policy}}$, and $\mathrm{C}(\Tilde{\sigma}, \Rewardvec^k)$ via $\Tilde{\policy}$ as in lines~\ref{alg:verif1}--\ref{alg:verif3} 
            \State \{$\hat{\policy} \gets \Tilde{\policy}$; $\trustregion \gets \trustregion \trustregion_0$\} if $\mathrm{C}(\Tilde{\sigma}, \Rewardvec^k) \geq \mathrm{C}(\hat{\sigma}, \Rewardvec^k)$ else \{$\trustregion \gets \trustregion / \trustregion_0$\} \Comment{Verification step}
        \EndWhile
    \State \Return $\policy^{k+1} := \hat{\policy}$
  \end{algorithmic}
\end{algorithm*}%

\paragraph{\myett{\textbf{Convergence to Local Optimum Solution}}}
\myett{ The convergence guarantees of the proposed sequential convex scheme with trust regions follow straightforwardly from the general convergence of sequential convex programming (SCP) schemes as proved in Theorem $3.14$ and Theorem $4.7$ of~\cite{mao2018successive}. Specifically, weak convergence is ensured as the SCP algorithm generates a set of convergent subsequences, all of which satisfy the first-order conditions~\cite{mao2018successive}. This is not convergence in its strict sense due to potential oscillation between several limit points. Still, surprisingly most of the convergence claims of nonlinear optimization schemes fall into this category. Furthermore, under the right regularity assumptions on the cost function, the authors of~\cite{mao2018successive} proved that SCP schemes with trust regions can converge to a local optimum solution with a superlinear convergence rate.
}
\section{Numerical Experiments}

We evaluate the proposed IRL algorithm on several POMDP instances from \cite{junges2020enforcing}, and a simulated wheeled ground robot operating in a high-fidelity, continuous, and 3-D Unity simulation.
We first compare our IRL algorithm with a straightforward variant of GAIL~\cite{ho2016generative} adapted for POMDPs.
Then, we provide results on the data-efficiency of the proposed approach when taking advantage of side information.
Finally, we demonstrate the scalability of the routine \texttt{SCPForward} for solving the \emph{forward} problem through comparisons with state-of-the-art solvers such as  $\texttt{SolvePOMDP}$~\cite{walraven2017accelerated}, $\texttt{SARSOP}$~\cite{kurniawati2008sarsop}, $\texttt{PRISM-POMDP}$~\cite{norman2017verification}. We provide the code for reproducibility of the results in this paper at https://github.com/wuwushrek/MCE\_IRL\_POMDPS.


\subsection{Simulation on Hand-Crafted POMDP Instances}
We first evaluate the proposed IRL algorithm on several POMDP instances extracted from the work \cite{junges2020enforcing}.

\begin{figure}[!hbt]
    \centering
    \definecolor{cheesebound}{rgb}{0.2, 0.2, 0.7}
\tikzset{cross/.style={cross out, draw, 
         minimum size=2*(#1-\pgflinewidth), 
         inner sep=0pt, outer sep=0pt}}
         
\begin{tikzpicture}[scale=0.6]
\draw[thin] (0,3) rectangle node {\color{black}1} (1,2);
\draw (1,3) rectangle node {\color{black}2} (2,2);
\draw (2,3) rectangle node {\color{black}3} (3,2);
\draw (3,3) rectangle node {\color{black}4} (4,2);
\draw (4,3) rectangle node {\color{black}5} (5,2);
\draw (0,2) rectangle node {\color{black}6} (1,1);
\draw (0,1) rectangle node {\color{black}9} (1,0);
\draw (0,0) rectangle node {\color{black}12} (1,-1); 
\draw (2,2) rectangle node {\color{black}7} (3,1);
\draw (2,1) rectangle node {\color{black}10} (3,0);
\draw (2,0) rectangle node {\color{black}13} (3,-1); 
\draw (4,2) rectangle node {\color{black}8} (5,1);
\draw (4,1) rectangle node {\color{black}11} (5,0);
\draw (4,0) rectangle node {\color{black}14} (5,-1); 

\draw (0.5, -0.5) node[cross=6pt, draw=red] {};
\draw [draw=green, very thick](2.5, -0.5) circle (10pt) {};
\draw (4.5, -0.5) node[cross=6pt, draw=yellow] {};

\draw[line width=0.5mm,cheesebound] (0,-1) edge (1,-1);
\draw[line width=0.5mm,cheesebound] (0,-1) edge (0,3);
\draw[line width=0.5mm,cheesebound] (5,3) edge (0,3);
\draw[line width=0.5mm,cheesebound] (1,2) edge (2,2);
\draw[line width=0.5mm,cheesebound] (1,2) edge (2,2);
\draw[line width=0.5mm,cheesebound] (3,2) edge (4,2);

\draw[line width=0.5mm,cheesebound] (2,-1) edge (3,-1);
\draw[line width=0.5mm,cheesebound] (4,-1) edge (5,-1);

\draw[line width=0.5mm,cheesebound] (5,-1) edge (5,3);

\draw[line width=0.5mm,cheesebound] (4,-1) edge (4,2);
\draw[line width=0.5mm,cheesebound] (3,-1) edge (3,2);
\draw[line width=0.5mm,cheesebound] (2,-1) edge (2,2);
\draw[line width=0.5mm,cheesebound] (1,-1) edge (1,2);

\end{tikzpicture}
    \frame{\includegraphics[scale=0.2]{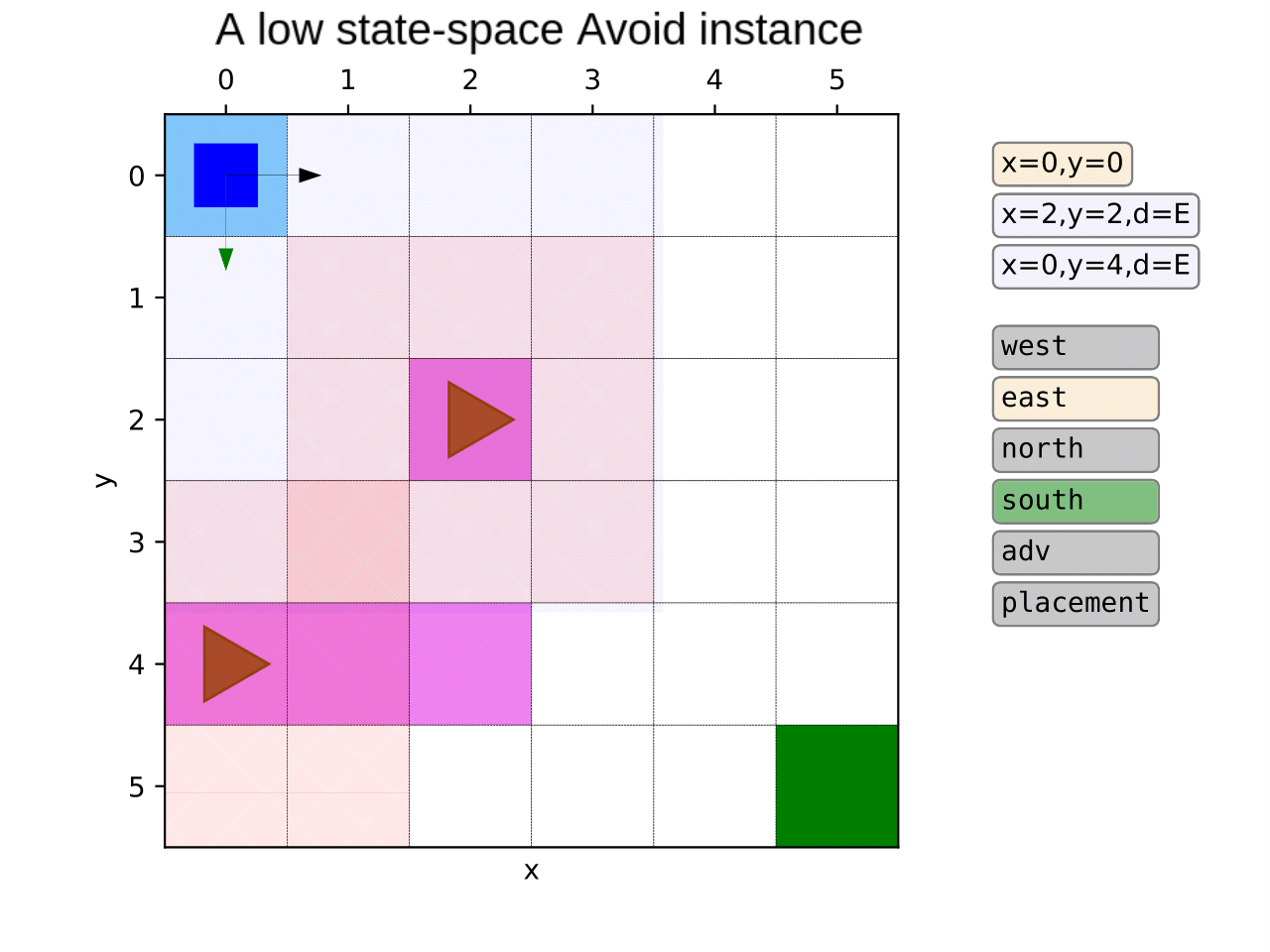}}
    \frame{\includegraphics[scale=0.2]{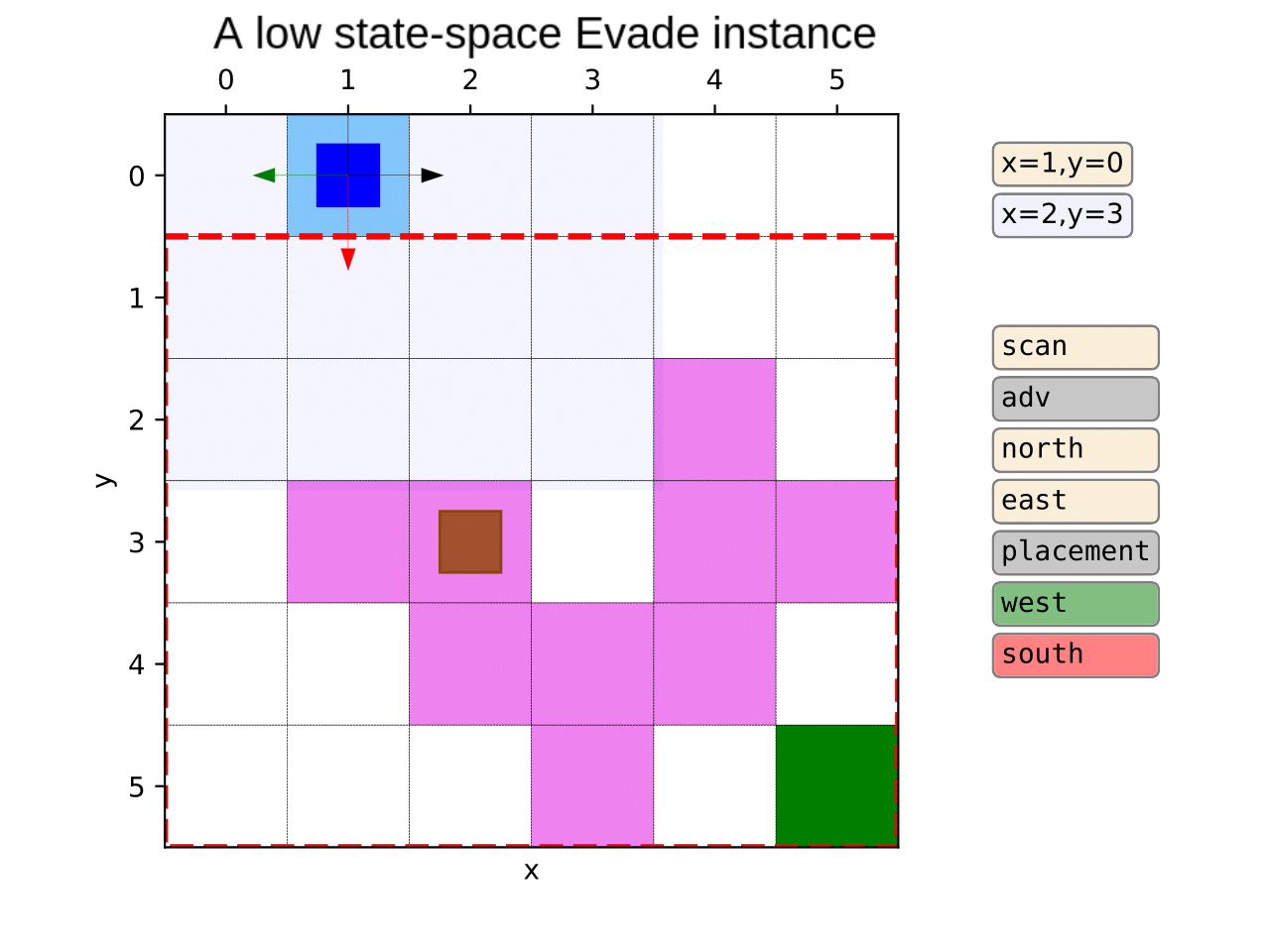}}
    \caption{Some examples from the benchmark set provided in~\cite{junges2020enforcing}. From left to right, we have the \emph{Maze}, \emph{Avoid}, and Evade environments, respectively.}
    \label{fig:benchmarkset}
\end{figure}
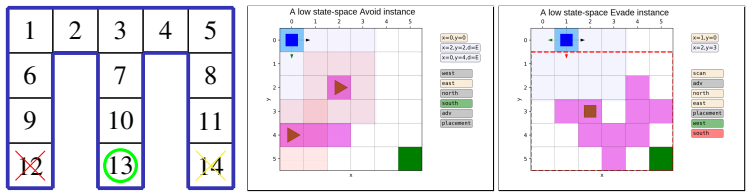

\paragraph{\textbf{Benchmark Set}} The POMDP instances are as follows.
$\emph{Evade}$ is a turn-based game where the agent must reach a destination without being intercepted by a faster player. 
In $\emph{Avoid}$, the agent must avoid being detected by two other moving players following certain preset, yet unknown routes.
In $\emph{Intercept}$, the agent must intercept another player who is trying to exit a gridworld. 
In $\emph{Rocks}$, the agents must sample at least one good rock over the several rocks without any failures.
In $\emph{Obstacle}$, an agent must find an exit in a gridworld without colliding with any static obstacles.
In these instances, the agent only observes a fixed radius around its current position, see Figure~\ref{fig:benchmarkset}.
Finally, in $\emph{Maze}$, the agent must exit a maze as fast as possible while observing only the walls around it and should not get stuck in any of the trap states.

\paragraph{\textbf{Variants of Learned Policies and Experts}}
We refer to four types of policies.
The type of policy depends on whether it uses side information from a temporal specification $\reachPropSymbol$ or not, and whether it uses a memory size $\mathrm{M}=1$ or $\mathrm{M}=10$.
We also consider two types of experts.
The first expert has full information about the environment and computes an optimal policy in the underlying MDP.
The second expert has partial observation and computes a locally optimal policy in the POMDP with a memory size of $\mathrm{M}=15$.
Recall that the agent always has partial information.
Therefore, the first type of expert corresponds to having information asymmetry between the learning agent and expert.
\emph{Besides, we consider as a baseline a variant of GAIL where we learn the policy on the MDP without side information, and extend it to POMDPs via an offline computation of the belief in the states. \myett{Specifically, we find the optimal policy on the MDP by solving the convex optimization problem corresponding to the forward problem on MDPs. The resulting policy is a state-based policy that needs to be transformed in order to act on a POMDP. The transformation is done by exploiting the expert demonstrations to construct a belief state. That is, the trajectories $\tau$ of the expert are used in a Bayesian belief updates~\eqref{eq:bayesian-update} to estimate the probability of being in each state of the POMDP. Thus, by combining the computed belief and the state-based policy, we obtain an observation-based policy for the POMDP.}
\myett{Doing so could provide a significant advantage to the GAIL variant since the state-based policy is the optimal policy on the MDP. However, despite the high performance in practice, the policy on the POMDP is generally suboptimal, even if the MDP policy were optimal.}
}

We discuss the effect of side information and memory in the corresponding policies.
While we detail only on the \emph{Maze} example, where the agent must exit a maze as fast as possible, we observe similar patterns for other examples.
Detailed results for the other examples are provided in the appendix.

\begin{figure*}[t]
\centering
    \begin{subfigure}[t]{\textwidth}
        \centering\hspace*{2.4cm}\begin{tikzpicture}
\definecolor{color1}{HTML}{FF0000}
\definecolor{color0}{HTML}{0000FF}
\definecolor{color2}{HTML}{FF00FF}
\begin{customlegend}[legend columns=3,legend style={
  fill opacity=0.8,
  draw opacity=1,
  text opacity=1,
  at={(1.3,0.4)},
  anchor=south,
  draw=white!80!black,
  scale=0.40,
    font=\titlesize,
  mark options={scale=0.5},
},legend entries={No information asymmetry, Under information asymmetry, GAIL}]

\addlegendimage{line width=2pt, color0}
\addlegendimage{line width=2pt, color1, dashed}
\addlegendimage{line width=2pt, color2, densely dashdotted}
\end{customlegend}
\end{tikzpicture}
    \end{subfigure}\hfill%
    \centering
    \begin{subfigure}[t]{0.541\textwidth}
        \centering
        \centering\captionsetup{width=.95\linewidth}%
\begin{tikzpicture}

\definecolor{color1}{HTML}{FF0000}
\definecolor{color0}{HTML}{0000FF}
\definecolor{color2}{HTML}{FF00FF}

\begin{axis}[
legend cell align={left},
legend columns=2,
legend style={
  fill opacity=0.8,
  draw opacity=1,
  text opacity=1,
  at={(1.15,1.25)},
  anchor=south,
    ticklabel style = {font=\tiny},
  draw=white!80!black,
  font=\fontsize{7.4}{7.4}\selectfont,
  scale=0.40
},
title={$R_\policy^\Rewardvec$},
title style={yshift=-1.5ex,xshift=-15ex,font=\titlesize},
width=6.cm,
height=3.1cm,
tick align=outside,
tick pos=left,
no markers,
every axis plot/.append style={line width=2pt},
x grid style={white!69.0196078431373!black},
xlabel={\footnotesize{Time steps}},
xlabel={\textcolor{gray!40!black}{\arial \textit{Finite-memory} policy}},
xlabel style={font=\titlesize,yshift=18ex},
xmajorgrids,
xmin=-4.95, xmax=103.95,
xtick style={color=black},
y grid style={white!69.0196078431373!black},
ylabel={\footnotesize{Mean  accumulated  reward}},
ylabel={\textcolor{gray!40!black}{\arial\textbf{Without} side} \\\textcolor{gray!40!black}{\arial information}},
ylabel style={rotate=-90,align=center,font=\titlesize},
xtick={0,25,50,75,100},
ymajorgrids,
ymin=-30.8979836098955, ymax=66.0315604206547,
ytick style={color=black}
]

\addplot[color0] table[x=x,y=y,mark=none] {maze_mem10_trajsize5pomdp_irl.tex};
\addplot[color1,dashed] table[x=x,y=y,mark=none] {maze_mem10_trajsize5mdp_irl.tex};


\addplot[color2,densely dashdotted] table[x=x,y=y,mark=none] {maze_mdp_fwd_gail.tex};
\addplot [name path=upper2,draw=none] table[x=x,y expr=\thisrow{y}+\thisrow{err}] {maze_mdp_fwd_gail.tex};
\addplot [name path=lower2,draw=none] table[x=x,y expr=\thisrow{y}-\thisrow{err}] {maze_mdp_fwd_gail.tex};
\addplot [fill=color2!40,opacity=0.5] fill between[of=upper2 and lower2];

\addplot [name path=upper1,draw=none] table[x=x,y expr=\thisrow{y}+\thisrow{err}]{maze_mem10_trajsize5mdp_irl.tex};
\addplot [name path=lower1,draw=none] table[x=x,y expr=\thisrow{y}-\thisrow{err}] {maze_mem10_trajsize5mdp_irl.tex};
\addplot [fill=color1!40,opacity=0.5] fill between[of=upper1 and lower1];

\addplot [name path=upper,draw=none] table[x=x,y expr=\thisrow{y}+\thisrow{err}] {maze_mem10_trajsize5pomdp_irl.tex};
\addplot [name path=lower,draw=none] table[x=x,y expr=\thisrow{y}-\thisrow{err}] {maze_mem10_trajsize5pomdp_irl.tex};
\addplot [fill=color0!40,opacity=0.5] fill between[of=upper and lower];

\end{axis}

\end{tikzpicture}
    \end{subfigure}%
    \begin{subfigure}[t]{0.541\textwidth}
        \centering
        \centering\captionsetup{width=.95\linewidth}%
\begin{tikzpicture}

\definecolor{color1}{HTML}{FF0000}
\definecolor{color0}{HTML}{0000FF}
\definecolor{color2}{HTML}{FF00FF}

\begin{axis}[
legend cell align={left},
legend columns=2,
legend style={
  fill opacity=0.8,
  draw opacity=1,
  text opacity=1,
  at={(0.5,1)},
  anchor=south,
  ticklabel style = {font=\tiny},
  draw=white!80!black,
  font=\fontsize{6}{6}\selectfont,
  scale=0.40
},
xshift=0.1cm,
width=6.cm,
height=3.1cm,
tick align=outside,
tick pos=left,
no markers,
every axis plot/.append style={line width=2pt},
x grid style={white!69.0196078431373!black},
xlabel={\footnotesize{Time steps}},
xlabel={},
xlabel={\textcolor{gray!40!black}{\arial \textit{Memoryless} policy}},
xlabel style={font=\titlesize,yshift=18ex},
xtick={0,25,50,75,100},
xmajorgrids,
xmin=-4.95, xmax=103.95,
xtick style={color=black},
y grid style={white!69.0196078431373!black},
ylabel={\footnotesize{Mean  accumulated  reward}},
ylabel={},
ymajorgrids,
ymin=-30.8979836098955, ymax=66.0315604206547,
ytick style={color=black},
ymajorticks=false,
]

\addplot[color0] table[x=x,y=y,mark=none] {maze_mem1_trajsize5pomdp_irl.tex};
\addplot[color1,dashed] table[x=x,y=y,mark=none] {maze_mem1_trajsize5mdp_irl.tex};


\addplot[color2,densely dashdotted] table[x=x,y=y,mark=none] {maze_mdp_fwd_gail.tex};
\addplot [name path=upper2,draw=none] table[x=x,y expr=\thisrow{y}+\thisrow{err}] {maze_mdp_fwd_gail.tex};
\addplot [name path=lower2,draw=none] table[x=x,y expr=\thisrow{y}-\thisrow{err}] {maze_mdp_fwd_gail.tex};
\addplot [fill=color2!40,opacity=0.5] fill between[of=upper2 and lower2];

\addplot [name path=upper1,draw=none] table[x=x,y expr=\thisrow{y}+\thisrow{err}]{maze_mem1_trajsize5mdp_irl.tex};
\addplot [name path=lower1,draw=none] table[x=x,y expr=\thisrow{y}-\thisrow{err}] {maze_mem1_trajsize5mdp_irl.tex};
\addplot [fill=color1!40,opacity=0.5] fill between[of=upper1 and lower1];

\addplot [name path=upper,draw=none] table[x=x,y expr=\thisrow{y}+\thisrow{err}] {maze_mem1_trajsize5pomdp_irl.tex};
\addplot [name path=lower,draw=none] table[x=x,y expr=\thisrow{y}-\thisrow{err}] {maze_mem1_trajsize5pomdp_irl.tex};
\addplot [fill=color0!40,opacity=0.5] fill between[of=upper and lower];

\end{axis}

\end{tikzpicture}
    \end{subfigure}%
    \vfill
    \centering
        \begin{subfigure}[t]{0.541\textwidth}
        \centering
        \centering\captionsetup{width=.95\linewidth}%
\begin{tikzpicture}

\definecolor{color1}{HTML}{FF0000}
\definecolor{color0}{HTML}{0000FF}
\definecolor{color2}{HTML}{FF00FF}
\begin{axis}[
legend cell align={left},
legend columns=2,
legend style={
  fill opacity=0.8,
  draw opacity=1,
  text opacity=1,
  at={(0.5,1)},
  anchor=south,
  ticklabel style = {font=\tiny},
  draw=white!80!black,
  font=\fontsize{6}{6}\selectfont,
  scale=0.40
},
title={\textbf{Finite-memory} policy \emph{with} side information},
title={$R_\policy^\Rewardvec$},
title style={yshift=-1.5ex,xshift=-15ex,font=\titlesize},
yshift=0.5cm,
width=6.cm,
height=3.1cm,
tick align=outside,
tick pos=left,
no markers,
every axis plot/.append style={line width=2pt},
x grid style={white!69.0196078431373!black},
xlabel={\titlesize{Time Steps}},
xmajorgrids,
xmin=-4.95, xmax=103.95,
xtick style={color=black},
y grid style={white!69.0196078431373!black},
xtick={0,25,50,75,100},
ylabel={\footnotesize{Mean accumulated  reward}},
ylabel={\titlesize{$R_\policy^\phi$}},
ylabel={\textcolor{gray!40!black}{\;\;\,\arial\textbf{With} side\;\;\,} \\\textcolor{gray!40!black}{\arial information}},
ylabel style={rotate=-90,align=center,font=\titlesize},
ymajorgrids,
ymin=-30.8979836098955, ymax=66.0315604206547,
ytick style={color=black}
]

\addplot[color0] table[x=x,y=y,mark=none] {maze_mem10_trajsize5pomdp_irl_si.tex};
\addplot[color1,dashed] table[x=x,y=y,mark=none] {maze_mem10_trajsize5mdp_irl_si.tex};


\addplot[color2,densely dashdotted] table[x=x,y=y,mark=none] {maze_mdp_fwd_gail.tex};
\addplot [name path=upper2,draw=none] table[x=x,y expr=\thisrow{y}+\thisrow{err}] {maze_mdp_fwd_gail.tex};
\addplot [name path=lower2,draw=none] table[x=x,y expr=\thisrow{y}-\thisrow{err}] {maze_mdp_fwd_gail.tex};
\addplot [fill=color2!40,opacity=0.5] fill between[of=upper2 and lower2];

\addplot [name path=upper1,draw=none] table[x=x,y expr=\thisrow{y}+\thisrow{err}]{maze_mem10_trajsize5mdp_irl_si.tex};
\addplot [name path=lower1,draw=none] table[x=x,y expr=\thisrow{y}-\thisrow{err}] {maze_mem10_trajsize5mdp_irl_si.tex};
\addplot [fill=color1!40,opacity=0.5] fill between[of=upper1 and lower1];

\addplot [name path=upper,draw=none] table[x=x,y expr=\thisrow{y}+\thisrow{err}] {maze_mem10_trajsize5pomdp_irl_si.tex};
\addplot [name path=lower,draw=none] table[x=x,y expr=\thisrow{y}-\thisrow{err}] {maze_mem10_trajsize5pomdp_irl_si.tex};
\addplot [fill=color0!40,opacity=0.5] fill between[of=upper and lower];

\end{axis}

\end{tikzpicture}
    \end{subfigure}%
    \begin{subfigure}[t]{0.541\textwidth}
        \centering
        \centering\captionsetup{width=.95\linewidth}%
        \vspace*{-3.03cm}
\begin{tikzpicture}

\definecolor{color1}{HTML}{FF0000}
\definecolor{color0}{HTML}{0000FF}
\definecolor{color2}{HTML}{FF00FF}
\begin{axis}[
legend cell align={left},
legend columns=2,
legend style={
  fill opacity=0.8,
  draw opacity=1,
  text opacity=1,
  at={(0.5,1)},
  anchor=south,
  ticklabel style = {font=\tiny},
  draw=white!80!black,
  font=\fontsize{6}{6}\selectfont,
  scale=0.40
},
xshift=0.1cm,
yshift = 0.0cm,
width=6.cm,
height=3.1cm,
tick align=outside,
tick pos=left,
no markers,
every axis plot/.append style={line width=2pt},
x grid style={white!69.0196078431373!black},
xlabel={\titlesize{Time Steps}},
xmajorgrids,
xmin=-4.95, xmax=103.95,
xtick style={color=black},
y grid style={white!69.0196078431373!black},
ylabel={\footnotesize{Mean  accumulated  reward}},
ylabel={},
ymajorgrids,
xtick={0,25,50,75,100},
ymin=-30.8979836098955, ymax=66.0315604206547,
ytick style={color=black},
ymajorticks=false,
]

\addplot[color0] table[x=x,y=y,mark=none] {maze_mem1_trajsize5pomdp_irl_si.tex};
\addplot[color1,dashed] table[x=x,y=y,mark=none] {maze_mem1_trajsize5mdp_irl_si.tex};


\addplot[color2,densely dashdotted] table[x=x,y=y,mark=none] {maze_mdp_fwd_gail.tex};
\addplot [name path=upper2,draw=none] table[x=x,y expr=\thisrow{y}+\thisrow{err}] {maze_mdp_fwd_gail.tex};
\addplot [name path=lower2,draw=none] table[x=x,y expr=\thisrow{y}-\thisrow{err}] {maze_mdp_fwd_gail.tex};
\addplot [fill=color2!40,opacity=0.5] fill between[of=upper2 and lower2];

\addplot [name path=upper1,draw=none] table[x=x,y expr=\thisrow{y}+\thisrow{err}]{maze_mem1_trajsize5mdp_irl_si.tex};
\addplot [name path=lower1,draw=none] table[x=x,y expr=\thisrow{y}-\thisrow{err}] {maze_mem1_trajsize5mdp_irl_si.tex};
\addplot [fill=color1!40,opacity=0.5] fill between[of=upper1 and lower1];

\addplot [name path=upper,draw=none] table[x=x,y expr=\thisrow{y}+\thisrow{err}] {maze_mem1_trajsize5pomdp_irl_si.tex};
\addplot [name path=lower,draw=none] table[x=x,y expr=\thisrow{y}-\thisrow{err}] {maze_mem1_trajsize5pomdp_irl_si.tex};
\addplot [fill=color0!40,opacity=0.5] fill between[of=upper and lower];

\end{axis}

\end{tikzpicture}
    \end{subfigure}%
    \vspace*{-0.8cm}
    \caption{
    Representative results on the $\mathrm{Maze}$ example; each sub-figure represents the average accumulated reward under the true reward function ($R_\policy^\Rewardvec$) over 1000 runs as a function of time.
  Compare the two rows: The policies in the top row that do not utilize side information suffer a performance drop under information asymmetry.
    On the other hand, in the bottom row, the performance of policies incorporating side information into learning does not decrease under information asymmetry.
    Compare the two columns: The performance of the finite-memory policies in the left column is significantly better than memoryless policies. Except for the memoryless policies without side information, our algorithm outperforms GAIL. The expert reward on the MDP is in average $48.22$, while we obtain the value $47.83$ for an expert acting on the POMDP.
    }
    \label{fig:maze}
\end{figure*}
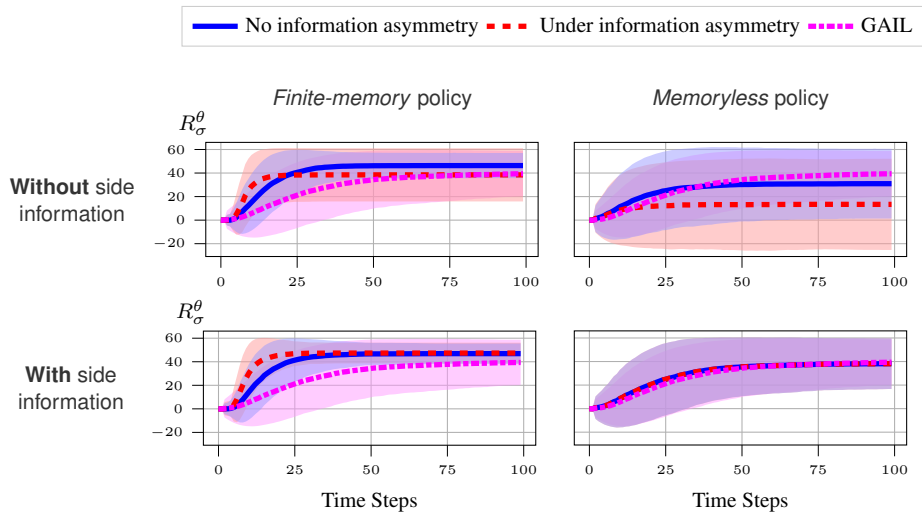%

\subsubsection{Maze Example}

The POMDP $\Pomdp$ is specified by $\States = \{s_1,\hdots,s_{14}\}$ corresponding to the cell labels in Figure~\ref{fig:benchmarkset}. An agent in the maze only observes whether or not there is a wall (in blue) in a neighboring cell. That is, the set of observations is $\Observations = \{o_1,\hdots,o_6, o_7\}$. For example, $o_1$ corresponds to observing west and north walls ($s_1$), $o_2$ to north and south walls ($s_2$, $s_4$), and $o_5$ to east and west walls ($s_6,s_7,s_8,s_9,s_{10},s_{11}$). The observations $o_6$ and $o_7$ denote the target state ($s_{13}$) and bad states($s_{12}$, $s_{14}$). The transition model is stochastic with a probability of slipping $p=0.1$. Further, the states $s_{13}$ and $s_{14}$ lead to the end of the simulation (trapping states).

In the IRL experiments, we consider three feature functions. We penalize taking more steps with $\phi^{\mathrm{time}}(\state,\action) = -1$ for all $\state,\action$. We provide a positive reward when reaching $s_{13}$ with $\phi^{\mathrm{target}}(\state,\alpha) = 1$ if $s = s_{13}$ and $\phi^{\mathrm{target}}(\state,\alpha) = 0$ otherwise. We penalize bad states $s_{12}$ and $s_{14}$ with $\phi^{\mathrm{bad}}(\state,\alpha) = -1$ if $s = s_{12}$ or $s = s_{14}$, and $\phi^{\mathrm{bad}}(\state,\alpha) = 0$ otherwise. \emph{Finally, we have the LTL formula $\reachPropSymbol = \textbf{G} \; \lnot \: \mathrm{bad}$ as the task specification, where $\mathrm{bad}$ is an atomic proposition that is true if the current state $\state = s_{12}$ or $\state = s_{14}$. We constrain the learned policy to satisfy $\mathrm{Pr}_{\Pomdp}^\policy(\textbf{G} \; \lnot \: \mathrm{bad}) \geq 0.9$.}

\paragraph{\textbf{Side Information Alleviates the Information Asymmetry}} Figure~\ref{fig:maze} shows that if there is an information asymmetry between the learning agent and the expert, the policies that do not utilize side information suffer a significant performance drop.
The policies that do not incorporate side information into learning obtain a lower performance by $57$\% under information asymmetry, as shown in the top row of Figure~\ref{fig:maze}.
On the other hand, as seen in the bottom row of Figure~\ref{fig:maze}, the performance of the policies that use side information is almost unaffected by the information asymmetry.

\begin{figure*}[t]
\centering
    \begin{subfigure}[t]{\textwidth}
        \centering\hspace*{1.4cm}\begin{tikzpicture}
\definecolor{color1}{HTML}{FF0000}
\definecolor{color0}{HTML}{0000FF}
\definecolor{color2}{HTML}{FF00FF}
\begin{customlegend}[legend columns=3,legend style={
  fill opacity=0.8,
  draw opacity=1,
  text opacity=1,
  at={(0.0,110.2)},
  anchor=south,
  draw=white!80!black,
  scale=0.40,
  mark options={scale=0.5},
},legend entries={With side information, Without side information, GAIL}]
\addlegendimage{line width=2pt, color0}
\addlegendimage{line width=2pt, color1,dashed}
\addlegendimage{line width=2pt, color2, densely dashdotted}
\end{customlegend}
\end{tikzpicture}
    \end{subfigure}\hfill%
    \vspace*{0.20cm}
    \centering
    \begin{subfigure}[t]{\textwidth}
        \centering
        \centering\captionsetup{width=.95\linewidth}%
\begin{tikzpicture}

\definecolor{color0}{HTML}{FF0000}
\definecolor{color1}{HTML}{0000FF}
\definecolor{color2}{HTML}{FF00FF}
\begin{axis}[
legend cell align={left},
legend columns=2,
legend style={
  fill opacity=0.8,
  draw opacity=1,
  text opacity=1,
  at={(0.5,1)},
  anchor=south,
  draw=white!80!black,
  font=\fontsize{6}{6}\selectfont,
  scale=0.40
},
title={$R_\policy^\phi$},
title={},
title style={yshift=-1.5ex,xshift=-15ex,font=\titlesize},
width=\textwidth,
height=0.3\textwidth,
tick align=outside,
tick pos=left,
no markers,
every axis plot/.append style={line width=2pt},
x grid style={white!69.0196078431373!black},
xlabel={{Time Steps}},
xmajorgrids,
xmin=-4.95, xmax=303.95,
xtick style={color=black},
y grid style={white!69.0196078431373!black},
ylabel={{Total Reward}},
ymajorgrids,
xtick={0,75,150,225,300},
ymin=-32.8979836098955, ymax=56.0315604206547,
ticklabel style = {font=\normalsize}
]

\addplot[color2,densely dashdotted] table[x=x,y=y,mark=none] {avoid_mdp_fwd_gail.tex};
\addplot [name path=upper2,draw=none] table[x=x,y expr=\thisrow{y}+\thisrow{err}] {avoid_mdp_fwd_gail.tex};
\addplot [name path=lower2,draw=none] table[x=x,y expr=\thisrow{y}-\thisrow{err}] {avoid_mdp_fwd_gail.tex};
\addplot [fill=color2!40,opacity=0.5] fill between[of=upper2 and lower2];

\addplot[color0,dashed] table[x=x,y=y,mark=none] {avoid_mem1_trajsize10mdp_irl.tex};
\addplot[color1] table[x=x,y=y,mark=none] {avoid_mem1_trajsize10mdp_irl_si.tex};

\addplot [name path=upper1,draw=none] table[x=x,y expr=\thisrow{y}+\thisrow{err}]{avoid_mem1_trajsize10mdp_irl_si.tex};
\addplot [name path=lower1,draw=none] table[x=x,y expr=\thisrow{y}-\thisrow{err}] {avoid_mem1_trajsize10mdp_irl_si.tex};
\addplot [fill=color1!40,opacity=0.5] fill between[of=upper1 and lower1];

\addplot [name path=upper,draw=none] table[x=x,y expr=\thisrow{y}+\thisrow{err}] {avoid_mem1_trajsize10mdp_irl.tex};
\addplot [name path=lower,draw=none] table[x=x,y expr=\thisrow{y}-\thisrow{err}] {avoid_mem1_trajsize10mdp_irl.tex};
\addplot [fill=color0!40,opacity=0.5] fill between[of=upper and lower];
\end{axis}

\end{tikzpicture}
    \end{subfigure}%
    \vspace*{-5mm}
    \caption{Representative results on the $\mathrm{Avoid}$ example showing the reward of the policies under the true reward function ($R_\policy^\Rewardvec$) versus the time steps.}
    \label{fig:avoid}
    \vspace*{-2mm}
\end{figure*}

\paragraph{\textbf{Memory Leads to More Performant Policies}}
The results in Figure~\ref{fig:maze} demonstrate that incorporating memory into the policies improves the performance, i.e., the attained reward, in all examples, both in solving the forward problem and learning policies from expert demonstrations.
Incorporating memory partially alleviates the effects of  information asymmetry, as the performance of the finite-memory policy decreases by $18$\% under information asymmetry as opposed to $57$\% for the memoryless policy.

We see that in Table~\ref{tab:ToolComp}, incorporating memory into policy on the $\mathrm{Maze}$ and $\mathrm{Rocks}$ benchmarks, allows $\texttt{SCPForward}$ to compute policies that are almost optimal, evidenced by obtaining almost the same reward as the solver $\texttt{SARSOP}$.

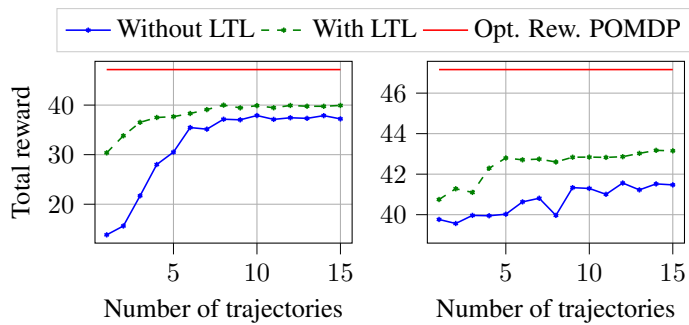
\begin{figure}[!hb]
    \centering
\begin{tikzpicture}

\begin{groupplot}[group style={group size=2 by 1}]
\nextgroupplot[
legend cell align={left},
legend columns=5,
legend style={
  fill opacity=0.8,
  draw opacity=1,
  text opacity=1,
  at={(-0.15,1.05)},
  anchor=south west,
  draw=white!80!black
},
width=5cm,
height=4cm,
tick align=outside,
tick pos=left,
x grid style={white!69.0196078431373!black},
xlabel={Number of trajectories},
xmajorgrids,
xmin=0.3, xmax=15.7,
xtick style={color=black},
y grid style={white!69.0196078431373!black},
ylabel={Total reward},
ymajorgrids,
ymin=12.15607, ymax=48.82853,
ytick style={color=black}
]
\addplot [semithick, blue, mark=asterisk, mark size=1, mark options={solid}]
table {%
1 13.8229999542236
2 15.6123332977295
3 21.7000007629395
4 28.0136661529541
5 30.5090007781982
6 35.466667175293
7 35.1426658630371
8 37.1469993591309
9 37.0320014953613
10 37.8953323364258
11 37.1166648864746
12 37.4440002441406
13 37.3230018615723
14 37.8726673126221
15 37.2173347473145
};
\addlegendentry{Without LTL}
\addplot [semithick, green!50.1960784313725!black, dashed, mark=asterisk, mark size=1, mark options={solid}]
table {%
1 30.3713340759277
2 33.8123321533203
3 36.5169982910156
4 37.4993324279785
5 37.666332244873
6 38.3046684265137
7 39.0913314819336
8 39.9926681518555
9 39.4426651000977
10 39.8916664123535
11 39.4599990844727
12 39.9243316650391
13 39.7676651000977
14 39.7566673278809
15 39.9330009460449
};
\addlegendentry{With LTL}
\addplot [semithick, red]
table {%
1 47.1615982055664
15 47.1615982055664
};
\addlegendentry{Opt. Rew. POMDP}

\nextgroupplot[
legend cell align={left},
legend columns=5,
legend style={
  fill opacity=0.8,
  draw opacity=1,
  text opacity=1,
  at={(0,1)},
  anchor=south west,
  draw=white!80!black
},
width=5cm,
height=4cm,
tick align=outside,
tick pos=left,
x grid style={white!69.0196078431373!black},
xlabel={Number of trajectories},
xmajorgrids,
xmin=0.3, xmax=15.7,
xtick style={color=black},
y grid style={white!69.0196078431373!black},
ymajorgrids,
ymin=38.59682, ymax=47.5694466666667,
ytick style={color=black}
]
\addplot [semithick, blue, mark=asterisk, mark size=1, mark options={solid}]
table {%
1 39.764331817627
2 39.564331817627
3 39.964331817627
4 39.9476661682129
5 40.0233345031738
6 40.6316680908203
7 40.8110008239746
8 39.964331817627
9 41.3343315124512
10 41.2949989318848
11 41.0046653747559
12 41.5623321533203
13 41.224666595459
14 41.5193328857422
15 41.4710006713867
};
\addplot [semithick, green!50.1960784313725!black, dashed, mark=asterisk, mark size=1, mark options={solid}]
table {%
1 40.7490005493164
2 41.2799987792969
3 41.099666595459
4 42.2910003662109
5 42.798999786377
6 42.7073348999023
7 42.7443321228027
8 42.601001739502
9 42.8359985351562
10 42.842000579834
11 42.8273323059082
12 42.8623344421387
13 43.02667388916
14 43.1749992370605
15 43.151668548584
};
\addplot [semithick, red, forget plot]
table {%
1 47.1615982055664
15 47.1615982055664
};
\end{groupplot}

\end{tikzpicture}
    \caption{We show the data efficiency of the proposed approach through the total reward obtained by the learned policies as a function of the number of expert demonstrations (No information asymmetry). The figure on the left shows the performance of learning memoryless policies, while the figure on the right shows the performance of a $5$-FSC.}
    \vspace*{-2mm}
    \label{fig:data-eff}
\end{figure}

\paragraph{\textbf{Side Information Improves Data Efficiency}}
Figure~\ref{fig:data-eff} shows that even on a low data regime, learning with task specifications achieves significantly better performance than without the task specifications. 

\paragraph{\textbf{Side Information Improves Performance}}
Besides, in a more complicated environment such as $\mathrm{Avoid}$, Figure~\ref{fig:avoid} shows that task specifications are crucial to hope even to learn the task. Specifically, $\mathrm{Avoid}[n,r,slip]$ is a turn-based game, where the agent must reach an exit point while avoiding being detected by two other moving players following certain predefined yet unknown routes. 
The agent can only observe the players if they are within a fixed radius from the agent's current position when the action \emph{scan} is performed.
Besides, with the players' speed being uncertain, their position in the routes can not be inferred by the agent. The parameters $n$, $r$, and $slip$ specify the dimension of the grid, the view radius, and the slippery probability, respectively.

We consider four feature functions to parameterize the unknown reward. The first feature provides a positive reward to the agent upon reaching the exit point.  The second feature penalizes the agent if it collides with a player.  The third feature penalizes the agent if it is detected by a player. The fourth feature imposes a penalty cost for each action taken. We encode the side information as the temporal logic task specification \emph{avoid being detected until reaching the exit point with probability greater than $0.98$}.

Figure~\ref{fig:avoid} shows that the algorithm is unable to learn without side information while side information induces a learned policy that is  optimal. Specifically, the learned policy without side information seems to only focus on avoiding being detected and collision as the corresponding learned features were close to zero.

\subsubsection{\texttt{SCPForward} Yields Better Scalability}
\begin{table*}[t]
	\setlength{\tabcolsep}{4.5pt}
	\centering
	\scalebox{0.8}{%
		\begin{tabular}{@{}cccc|cc|cc|cc@{}}
			\toprule
			&  & & & \multicolumn{2}{c|}{$\texttt{SCPForward}$}   &\multicolumn{2}{c|}{$\texttt{SARSOP}$} & \multicolumn{2}{c}{$\texttt{SolvePOMDP}$}  \\
			Problem   & $|\States|$ & $|\States \times \Observations|$ & $|\Observations|$ & $R_\policy^\Rewardvec$ & Time (s) & $R_\policy^\Rewardvec$ & Time (s) & $R_\policy^\Rewardvec$ & Time (s) \\
			\midrule
			$\mathrm{Maze}$ & $17$  & $162$ & $11$ & $39.24$ &   $\mathbf{0.1}$ & $\mathbf{47.83}$ & $0.24$ &  $47.83$ & $0.33$\\
			$\mathrm{Maze}$ ($3$-$\mathrm{FSC})$ & $49$  & $777$ & $31$ & $44.98$ &   $\mathbf{0.6}$ & NA & NA &  NA & NA\\
			$\mathrm{Maze}$ ($10$-$\mathrm{FSC})$ & $161$  & $2891$ & $101$ & $46.32$ &   $2.04$ & NA & NA &  NA & NA\\
			$\mathrm{Obstacle}[10]$ & $102$  & $1126$ & $5$ & $19.71$ &   $8.79$ & $\mathbf{19.8}$ & $\mathbf{0.02}$ & $5.05$ & $3600$ \\
			$\mathrm{Obstacle}[10]$($5$-$\mathrm{FSC})$ & $679$  & $7545$ & $31$ & $19.77$ &   $38$ & NA & NA & NA & NA \\
			$\mathrm{Obstacle}[25]$ & $627$  & $7306$ & $5$ & $19.59$ &   $14.22$ & $\textbf{19.8}$ & $\textbf{0.1}$ & $5.05$ & $3600$ \\
			$\mathrm{Rock}$ & $550$  & $4643$ & $67$ & $19.68$ &   $12.2$ & $\mathbf{19.83}$ & $\mathbf{0.05}$ &  $-$ & $-$\\
			$\mathrm{Rock}$ ($3$-$\mathrm{FSC})$ & $1648$  & $23203$ & $199$ & $19.8$ &   $15.25$ & NA & NA &  $-$ & $-$\\
			$\mathrm{Rock}$ ($5$-$\mathrm{FSC})$ & $2746$  & $41759$ & $331$ & $19.82$ &   $97.84$ & NA & NA &  $-$ & $-$\\
			$\mathrm{Intercept}[5,2,0]$ & $1321$  & $5021$ & $1025$ & $\mathbf{19.83}$ &   $\mathbf{10.28}$ & $\mathbf{19.83}$ & $13.71$ &  $-$ & $-$\\
			$\mathrm{Intercept}[5,2,0.1]$ & $1321$  & $7041$ & $1025$ & $\mathbf{19.81}$ &   $\mathbf{13.18}$ & $\mathbf{19.81}$ & $81.19$ &  $-$ & $-$\\
			$\mathrm{Evade}[5,2,0]$ & $2081$  & $13561$ & $1089$ & $\mathbf{97.3}$ &   $\mathbf{26.25}$ & $\mathbf{97.3}$ & $3600$ &  $-$ & $-$\\
			$\mathrm{Evade}[5,2,0.1]$ & $2081$  & $16761$ & $1089$ & $\mathbf{96.79}$ &   $\mathbf{26.25}$ & $95.28$ & $3600$ &  $-$ & $-$\\
			$\mathrm{Evade}[10,2,0]$ & $36361$  & $341121$ & $18383$ & $\mathbf{94.97}$ &   $\mathbf{3600}$ & $-$ & $-$ &  $-$ & $-$\\
			$\mathrm{Avoid}[4,2,0]$ & $2241$  & $5697$ & $1956$ & $\mathbf{9.86}$ &   $34.74$ & $\mathbf{9.86}$ & $\mathbf{9.19}$ &  $-$ & $-$\\
			$\mathrm{Avoid}[4,2,0.1]$ & $2241$  & $8833$ & $1956$ & $\mathbf{9.86}$ &   $\mathbf{14.63}$ & $\mathbf{9.86}$ & $210.47$ &  $-$ & $-$\\
			$\mathrm{Avoid}[7,2,0]$ & $19797$  & $62133$ & $3164$ & $\mathbf{9.72}$ &   $\mathbf{3503}$ & $-$ & $-$ &  $-$ & $-$\\
			\bottomrule
	\end{tabular}}
	\caption{Results for the benchmark sets for solving the forward problem. On larger benchmarks (e.g., $\mathrm{Evade}$ and $\mathrm{Avoid}$), $\texttt{SCPForward}$ can compute locally optimal policies, while the other solvers fail to provide solutions in the given time limit. In the environments $\mathrm{Obstacle}[n]$, $\mathrm{Intercept}[n,r,\mathrm{slip}]$, $\mathrm{Evade}[n,r,\mathrm{slip}]$, and $\mathrm{Avoid}[n,r,\mathrm{slip}]$, the parameters $n$, $r$, and $\mathrm{slip}$ are the size of the gridworld,  the view radius of the agent, and the probability of slippery, respectively.
	We set the time-out to $3600$ seconds. An empty cell (denoted by $-$) represents the solver failed to compute any policy before the time-out, while NA refers to not applicable due to the approach being based on belief updates.}
	\label{tab:ToolComp}
\end{table*}
We highlight three observations regarding the scalability of $\texttt{SCPForward}$.
First, the results in Table~\ref{tab:ToolComp} show that only $\texttt{SARSOP}$ is competitive with $\texttt{SCPForward}$ on larger POMDPs.
$\texttt{SolvePOMDP}$ runs out of time in all but the smallest benchmarks, and $\texttt{PrismPOMDP}$ runs out of memory in all benchmarks. Most of these approaches are based on updating a belief over the states, which for a large state space can become extremely computationally expensive.

Second, in the benchmarks with smaller state spaces, e.g., \emph{Maze} and \emph{Rock}, $\texttt{SARSOP}$ can compute policies that yield better performance in less time. This is due to the efficiency of belief-based approaches on small-size problems.
On the other hand, $\texttt{SARSOP}$ does not scale to larger POMDPs with a larger number of states and observations.
For example, by increasing the number of transitions in \emph{Intercept} benchmark from $5021$ to $7041$, the computation time for $\texttt{SARSOP}$ increases by $516$\%.
On the other hand, the increase of the computation time of $\texttt{SCPForward}$ is only $28$\%.

Third, on the largest benchmarks, including tens of thousands of states and observations, $\texttt{SARSOP}$ fails to compute any policy before time-out, while $\texttt{SCPForward}$ found a solution.
Finally, we also note that $\texttt{SCPForward}$ can also compute a policy that maximizes the causal entropy and satisfies an LTL specification, unlike $\texttt{SARSOP}$.
\begin{figure*}[t]
\centerline{
    \includegraphics[scale=0.185]{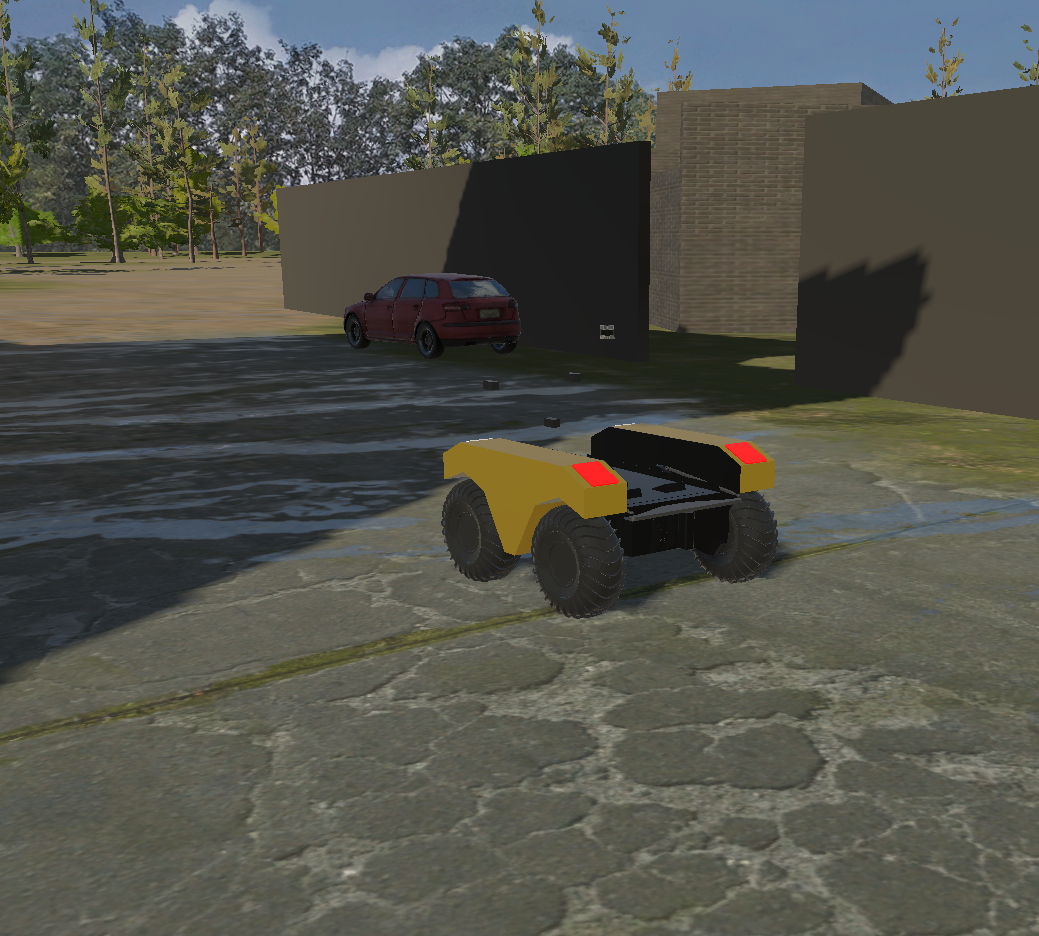}
    \includegraphics[scale=0.416]{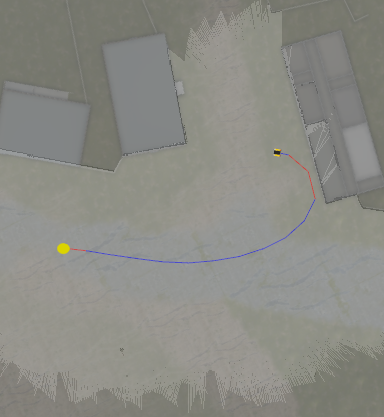}
}
\caption{Left: A simulated Clearpath Warthog operating in a Unity simulation. Right: A demonstration provided by an expert.}
\label{fig_simulated_warty}
\end{figure*}
\subsection{Simulation on a Ground Robot}
\begin{figure*}[t]
    \centering
    \input{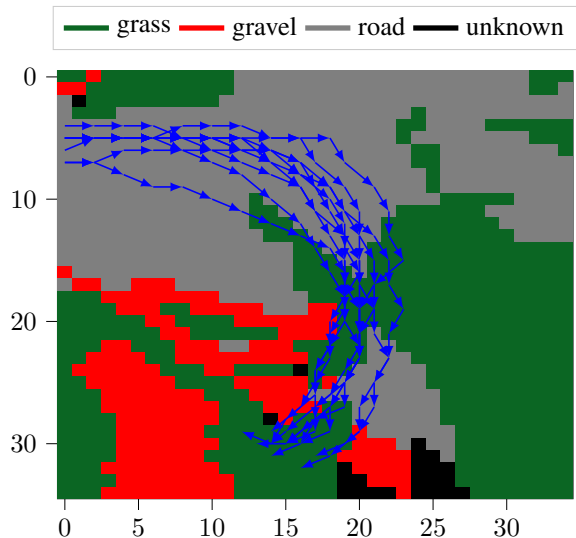}
    \caption{Gridworld representation of the environment. The figure shows the area of the unity environment where we applied the developed algorithm.}
    \label{fig:gridphoenix}
\end{figure*}

We demonstrate an application of the proposed algorithm in a continuous 3-D Unity environment containing a ClearPath warthog operating in a semi-structured village.
A screen shot of the robot operating in this environment and its corresponding trajectory can be seen in Figure \ref{fig_simulated_warty}.
This environment contains a variety of obstacles including buildings, trees, and vehicles as well as three terrain types describing our features, $\phi$, grass, gravel, and road.
The simulated environment operates in a  state space consisting of $3350$ states, $33254$ transitions and $944$ total observations.
This simulation is used to gather data for training, and test an agent's ability to follow a policy from the learned reward function in two experimental scenarios.
\myett{In this experiment, we demonstrate the agent's ability to learn a reward function from demonstrations that are sub-optimal with respect to a known, true reward function. We also show how the learned policies perform compared to the optimal policies with full and partial observations obtained by solving the MDP or POMDP problem with the true reward function.}

The ground vehicle contains an autonomy stack consisting of three main subsystems\textemdash mapping, perception, and planning.
The mapping subsystem based on Omni-Mapper\cite{Trevor14ICRA} performs simultaneous localization and mapping (SLAM) using LiDAR and IMU sensors, providing a map used during planning.
The perception subsystem provides pixel level semantic segmentation for each image in a video stream from a RGB camera to an ontology of terrain and object classes.
Each semantic image is passed to a terrain projection algorithm which builds $N$ binary occupancy feature maps of the known environment used for reward learning where $N$ is the number of features.
The planning subsystem uses the maps produced from previous subsystems and the trajectory from a learned policy to autonomously navigate to a waypoint. 

\paragraph{\textbf{Expert Demonstrations and Reward Feature Encoding}}
We collected $10$ demonstrations of an expert teleoperating a robot to a predetermined waypoint (see Figure~\ref{fig:gridphoenix}).
The expert has an implicit preference to traverse the road followed by grass, and lastly gravel.
Consequently, we encode the unknown reward function as a linear combination of known features: $\Reward^\Rewardvec = \Rewardvec_1 \phi^{\mathrm{road}} + \Rewardvec_2 \phi^{\mathrm{gravel}} + \Rewardvec_3 \phi^{\mathrm{grass}} + \Rewardvec_4 \phi^{\mathrm{time}} + \Rewardvec_5 \phi^{\mathrm{goal}}$, where $\phi^{i}$ returns a value of $0$ when the feature of the corresponding state is not feature $i$, or $1$ otherwise.
In order to incentivize the shortest path, the feature $\mathrm{time}$ penalizes the number of actions taken in the environment before reaching the waypoint.
Furthermore, $\mathrm{goal}$ provides a positive reward upon reaching the waypoint.
For comparisons of the learned policy, we use the values $\Rewardvec=[0.2, -30, -2, -0.5, 50]$ as the ground truth reward weight vector. \myett{We emphasize that the demonstrations are sub-optimal with respect to the above ground truth reward as the vehicle often traverses gravel, corresponding to a high penalty reward.}

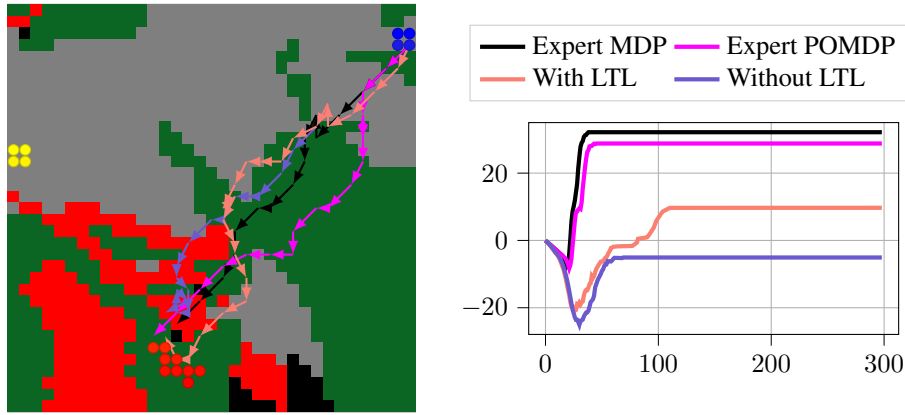
\begin{figure}
    \centering
    \begin{subfigure}{.48\textwidth}
        \centering
\begin{tikzpicture}

\definecolor{color0}{rgb}{0,1,1}
\definecolor{color1}{rgb}{1,1,0}
\definecolor{color2}{rgb}{1,0,1}
\definecolor{color3}{rgb}{0.415686274509804,0.352941176470588,0.803921568627451}
\definecolor{color4}{rgb}{0.980392156862745,0.501960784313725,0.447058823529412}

\begin{axis}[
width=7cm,
height=7cm,
tick align=outside,
tick pos=left,
x grid style={white!69.0196078431373!black},
xmin=-0.5, xmax=34.5,
xtick style={color=black},
y dir=reverse,
y grid style={white!69.0196078431373!black},
ymin=-0.5, ymax=34.5,
ytick style={color=black},
ymajorticks=false,
xmajorticks=false
]
\addplot [draw=color0, fill=color0, mark=*, only marks, scatter]
table{%
x  y
33 2
33 3
34 2
34 3
0 12
0 13
1 12
1 13
};
\addplot [draw=color1, fill=color1, mark=*, only marks, scatter]
table{%
x  y
13 30
15 32
12 29
16 31
15 31
14 31
14 31
13 31
13 29
14 30
};
\addplot graphics [includegraphics cmd=\pgfimage,xmin=-0.5, xmax=34.5, ymin=34.5, ymax=-0.5] {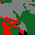};
\draw[-latex,thick,draw=black] (axis cs:34,3) -- (axis cs:32,5);
\draw[-latex,thick,draw=black] (axis cs:32,5) -- (axis cs:30,7);
\draw[-latex,thick,draw=black] (axis cs:30,7) -- (axis cs:28,9);
\draw[-latex,thick,draw=black] (axis cs:28,9) -- (axis cs:26,11);
\draw[-latex,thick,draw=black] (axis cs:26,11) -- (axis cs:26.0037634408602,9);
\draw[-latex,thick,draw=black] (axis cs:26,9) -- (axis cs:25,11);
\draw[-latex,thick,draw=black] (axis cs:25,11) -- (axis cs:25.0037634408602,13);
\draw[-latex,thick,draw=black] (axis cs:25,13) -- (axis cs:24,15);
\draw[-latex,thick,draw=black] (axis cs:24,15) -- (axis cs:22,17);
\draw[-latex,thick,draw=black] (axis cs:22,17) -- (axis cs:21,17);
\draw[-latex,thick,draw=black] (axis cs:21,17) -- (axis cs:19,19);
\draw[-latex,thick,draw=black] (axis cs:19,19) -- (axis cs:19.0037634408602,21);
\draw[-latex,thick,draw=black] (axis cs:19,21) -- (axis cs:18,23);
\draw[-latex,thick,draw=black] (axis cs:18,23) -- (axis cs:16,25);
\draw[-latex,thick,draw=black] (axis cs:16,25) -- (axis cs:14,27);
\draw[-latex,thick,draw=color2] (axis cs:34,3) -- (axis cs:32,5);
\draw[-latex,thick,draw=color2] (axis cs:32,5) -- (axis cs:30,7);
\draw[-latex,thick,draw=color2] (axis cs:30,7) -- (axis cs:30.0037634408602,9);
\draw[-latex,thick,draw=color2] (axis cs:30,9) -- (axis cs:30.0037634408602,11);
\draw[-latex,thick,draw=color2] (axis cs:30,11) -- (axis cs:30.0037634408602,9);
\draw[-latex,thick,draw=color2] (axis cs:30,9) -- (axis cs:30.0037634408602,11);
\draw[-latex,thick,draw=color2] (axis cs:30,11) -- (axis cs:30.0037634408602,13);
\draw[-latex,thick,draw=color2] (axis cs:30,13) -- (axis cs:29,15);
\draw[-latex,thick,draw=color2] (axis cs:29,15) -- (axis cs:27,17);
\draw[-latex,thick,draw=color2] (axis cs:27,17) -- (axis cs:26,17);
\draw[-latex,thick,draw=color2] (axis cs:26,17) -- (axis cs:24,19);
\draw[-latex,thick,draw=color2] (axis cs:24,19) -- (axis cs:24.0037634408602,21);
\draw[-latex,thick,draw=color2] (axis cs:24,21) -- (axis cs:22,21);
\draw[-latex,thick,draw=color2] (axis cs:22,21) -- (axis cs:20,21);
\draw[-latex,thick,draw=color2] (axis cs:20,21) -- (axis cs:18,22);
\draw[-latex,thick,draw=color2] (axis cs:18,22) -- (axis cs:16,24);
\draw[-latex,thick,draw=color2] (axis cs:16,24) -- (axis cs:14,26);
\draw[-latex,thick,draw=color2] (axis cs:14,26) -- (axis cs:12,28);
\draw[-latex,thick,draw=color3] (axis cs:34,3) -- (axis cs:33,5);
\draw[-latex,thick,draw=color3] (axis cs:33,5) -- (axis cs:31,7);
\draw[-latex,thick,draw=color3] (axis cs:31,7) -- (axis cs:29,9);
\draw[-latex,thick,draw=color3] (axis cs:29,9) -- (axis cs:27,10);
\draw[-latex,thick,draw=color3] (axis cs:27,10) -- (axis cs:27.0037634408602,8);
\draw[-latex,thick,draw=color3] (axis cs:27,8) -- (axis cs:26,10);
\draw[-latex,thick,draw=color3] (axis cs:26,10) -- (axis cs:24,12);
\draw[-latex,thick,draw=color3] (axis cs:24,12) -- (axis cs:23,14);
\draw[-latex,thick,draw=color3] (axis cs:23,14) -- (axis cs:21,16);
\draw[-latex,thick,draw=color3] (axis cs:21,16) -- (axis cs:20,16);
\draw[-latex,thick,draw=color3] (axis cs:20,16) -- (axis cs:19,16);
\draw[-latex,thick,draw=color3] (axis cs:19,16) -- (axis cs:18,18);
\draw[-latex,thick,draw=color3] (axis cs:18,18) -- (axis cs:17,18);
\draw[-latex,thick,draw=color3] (axis cs:17,18) -- (axis cs:15,20);
\draw[-latex,thick,draw=color3] (axis cs:15,20) -- (axis cs:14,22);
\draw[-latex,thick,draw=color3] (axis cs:14,22) -- (axis cs:14.0037634408602,23);
\draw[-latex,thick,draw=color3] (axis cs:14,23) -- (axis cs:15,24);
\draw[-latex,thick,draw=color3] (axis cs:15,24) -- (axis cs:15.0037634408602,26);
\draw[-latex,thick,draw=color3] (axis cs:15,26) -- (axis cs:14,26);
\draw[-latex,thick,draw=color3] (axis cs:14,26) -- (axis cs:14.0037634408602,24);
\draw[-latex,thick,draw=color3] (axis cs:14,24) -- (axis cs:15,25);
\draw[-latex,thick,draw=color3] (axis cs:15,25) -- (axis cs:13,26);
\draw[-latex,thick,draw=color4] (axis cs:34,3) -- (axis cs:33,5);
\draw[-latex,thick,draw=color4] (axis cs:33,5) -- (axis cs:31,7);
\draw[-latex,thick,draw=color4] (axis cs:31,7) -- (axis cs:29,9);
\draw[-latex,thick,draw=color4] (axis cs:29,9) -- (axis cs:27,10);
\draw[-latex,thick,draw=color4] (axis cs:27,10) -- (axis cs:27.0037634408602,8);
\draw[-latex,thick,draw=color4] (axis cs:27,8) -- (axis cs:26,10);
\draw[-latex,thick,draw=color4] (axis cs:26,10) -- (axis cs:24,11);
\draw[-latex,thick,draw=color4] (axis cs:24,11) -- (axis cs:23,13);
\draw[-latex,thick,draw=color4] (axis cs:23,13) -- (axis cs:21,13);
\draw[-latex,thick,draw=color4] (axis cs:21,13) -- (axis cs:20,13);
\draw[-latex,thick,draw=color4] (axis cs:20,13) -- (axis cs:19,15);
\draw[-latex,thick,draw=color4] (axis cs:19,15) -- (axis cs:18,17);
\draw[-latex,thick,draw=color4] (axis cs:18,17) -- (axis cs:19,16);
\draw[-latex,thick,draw=color4] (axis cs:19,16) -- (axis cs:18,18);
\draw[-latex,thick,draw=color4] (axis cs:18,18) -- (axis cs:19,20);
\draw[-latex,thick,draw=color4] (axis cs:19,20) -- (axis cs:19.0037634408602,21);
\draw[-latex,thick,draw=color4] (axis cs:19,21) -- (axis cs:20,23);
\draw[-latex,thick,draw=color4] (axis cs:20,23) -- (axis cs:20.0037634408602,25);
\draw[-latex,thick,draw=color4] (axis cs:20,25) -- (axis cs:18,26);
\draw[-latex,thick,draw=color4] (axis cs:18,26) -- (axis cs:16,28);
\draw[-latex,thick,draw=color4] (axis cs:16,28) -- (axis cs:15,30);
\draw[-latex,thick,draw=color4] (axis cs:15,30) -- (axis cs:14,30);
\draw[-latex,thick,draw=color4] (axis cs:14,30) -- (axis cs:13,28);
\end{axis}

\end{tikzpicture}
        \caption{The trajectories resulting from executing each policy with and without task specifications. The learner exploiting task specifications (orange) is able to reach one of the target states, while avoiding the gravel along the path. In contrast, the learner without side information (purple) fails to avoid the gravel.}
        \label{fig:traj_res}
    \end{subfigure}\
    \hspace{0.2em}
    \begin{subfigure}{.48\textwidth}
        \centering
\begin{tikzpicture}

\definecolor{color0}{rgb}{1,0,1}
\definecolor{color1}{rgb}{0.415686274509804,0.352941176470588,0.803921568627451}
\definecolor{color2}{rgb}{0.980392156862745,0.501960784313725,0.447058823529412}

\begin{axis}[
legend cell align={left},
legend columns=2,
legend style={
  fill opacity=0.8,
  draw opacity=1,
  text opacity=1,
  at={(-0.16,1.1)},
  anchor=south west,
  draw=white!80!black
},
width=6.5cm,
height=4.4cm,
tick align=outside,
tick pos=left,
x grid style={white!69.0196078431373!black},
xmajorgrids,
xmin=-14.9, xmax=312.9,
xtick style={color=black},
y grid style={white!69.0196078431373!black},
ymajorgrids,
ymin=-27.9316166666667, ymax=35.0166166666667,
ytick style={color=black}
]
\addplot [ultra thick, black]
table {%
0 0
15 -4.50733327865601
16 -5.010666847229
17 -5.66266679763794
18 -6.60799980163574
19 -7.79533338546753
20 -7.39400005340576
21 -3.98566675186157
22 0.419999957084656
23 5.57966661453247
24 8.59566688537598
25 10.1393337249756
26 11.8803329467773
28 16.742000579834
29 21.3756675720215
30 24.8530006408691
31 27.7999992370605
32 28.9729995727539
33 29.5053329467773
34 30.7636661529541
35 31.2136669158936
36 31.3386669158936
37 31.82200050354
38 32.1553344726562
298 32.1553344726562
};
\addlegendentry{Expert MDP}
\addplot [ultra thick, color0]
table {%
0 0
17 -5.11866664886475
18 -5.55200004577637
19 -6.29866647720337
20 -7.19999980926514
21 -8.31400012969971
22 -7.30700016021729
23 -5.3973331451416
24 -1.52799999713898
26 5.7736668586731
27 8.39833354949951
28 8.57833290100098
29 9.14266681671143
30 9.5116662979126
31 9.31966686248779
32 11.3070001602173
33 15.4333333969116
35 21.683666229248
36 24.2733325958252
37 26.2973327636719
38 26.8473339080811
39 27.9556674957275
40 28.0723323822021
41 28.1040000915527
43 28.7373332977295
44 28.6356658935547
45 28.8023338317871
298 28.8023338317871
};
\addlegendentry{Expert POMDP}
\addplot [ultra thick, color2]
table {%
0 0
1 -0.299999952316284
2 -0.636666655540466
3 -1.03199994564056
4 -1.52066671848297
5 -1.93799996376038
6 -2.39933323860168
7 -2.81666660308838
8 -3.34200000762939
11 -4.62333345413208
12 -5.21466684341431
13 -5.92133331298828
15 -7.81400012969971
16 -8.67533302307129
18 -11.4759998321533
19 -13.0880002975464
20 -14.7506666183472
21 -16.2433338165283
22 -18.3540000915527
23 -19.595666885376
24 -19.6226673126221
25 -20.9303340911865
26 -20.8466663360596
27 -20.6843338012695
28 -18.9720001220703
29 -19.2146663665771
30 -18.4653339385986
31 -19.011999130249
32 -18.4166660308838
33 -18.1520004272461
34 -16.3813323974609
35 -15.1966667175293
36 -14.7106666564941
37 -14.7773332595825
38 -14.0853328704834
39 -12.9753332138062
40 -12.7416667938232
41 -10.6269998550415
42 -11.1646671295166
43 -10.6573333740234
44 -9.39166641235352
45 -8.47166633605957
46 -7.91433334350586
47 -7.30700016021729
48 -6.82466650009155
49 -6.30066680908203
50 -6.26733350753784
51 -6.135666847229
52 -5.62900018692017
53 -5.16233348846436
54 -5.01233339309692
55 -4.32066679000854
56 -3.58733344078064
57 -3.11233339309692
58 -2.25399994850159
59 -2.1289999961853
60 -1.8900009155273
64 -1.75566674232483
66 -1.73233325004578
67 -1.71066659927368
70 -1.7056666469574
71 -1.69566655158997
76 -1.64900004386902
77 -1.62066674232483
81 -0.65400004386902
82 0.48733329772949
90 0.91233325004578
92 1.77066659927368
96 3.8456666469574
98 5.69566655158997
102 7.77900004386902
104 8.62066674232483
110 9.65400004386902
115 9.68733329772949
298 9.69733329772949
};
\addlegendentry{With LTL}
\addplot [ultra thick, color1]
table {%
0 0
2 -0.621999979019165
3 -0.980666637420654
5 -1.7756667137146
7 -2.69433331489563
8 -3.11700010299683
9 -3.56900000572205
10 -3.95499992370605
11 -4.37766647338867
12 -4.86633348464966
13 -5.45766687393188
14 -6.19466686248779
15 -6.98299980163574
16 -7.93433332443237
17 -9.05566692352295
18 -10.2443332672119
19 -11.5989999771118
20 -13.2163333892822
21 -14.9636669158936
22 -16.8403339385986
23 -18.7730007171631
24 -20.5833339691162
25 -21.6420001983643
26 -22.8793334960938
27 -23.5926666259766
28 -23.5769996643066
29 -24.2290000915527
30 -25.070333480835
31 -23.8333339691162
32 -23.7770004272461
33 -23.5249996185303
34 -22.5863342285156
35 -21.9636669158936
36 -21.4376659393311
37 -21.3796672821045
38 -20.9050006866455
39 -19.6749992370605
40 -18.3296661376953
41 -17.9446659088135
42 -17.4860000610352
43 -16.4629993438721
44 -13.933666229248
45 -12.9876670837402
46 -11.9886665344238
47 -11.2489995956421
48 -10.6113328933716
49 -9.94566631317139
50 -9.04666709899902
51 -8.62166690826416
52 -7.63833332061768
53 -7.121666431427
54 -7.09666681289673
55 -6.17999982833862
56 -6.26333332061768
57 -6.17166662216187
58 -6.246666431427
59 -5.97166681289673
60 -5.50500011444092
61 -5.17999982833862
68 -5.23833322525024
69 -5.0716667175293
298 -5.0716667175293
};
\addlegendentry{Without LTL}
\end{axis}

\end{tikzpicture}
        \caption{Evolution of the cumulative reward obtained by the learner as a function of the number of environment interactions. \myett{Expert MDP and Expert POMDP are the optimal policies on the MDP and POMDP, respectively for the ground truth reward function.}}
        \label{fig:rew-evolution}
    \end{subfigure}
    \caption{Impact of incorporating task specifications into reward learning.}
\end{figure}

\paragraph{\textbf{Modeling Robot Dynamics as POMDPs}}
From a ground truth map of the environment in the simulation, we obtain a high-level MDP abstraction of the learner's behavior on the entire state space.
Then, we impose a partial observability of the robot as follows: The robot does not see the entire map of the world but only see a fixed radius $r=4$ (in terms of the number of grid cells) around its current position.
Furthermore, we also incorporate uncertainty on the sensor classification of terrain features such that with probability $p=0.9$ the prediction is correct.

\paragraph{\textbf{Task Specifications}}
In addition to the expert demonstrations, we constrain the learned policy to satisfy $\mathrm{Pr}_{\Pomdp}^\policy(\lnot \: \mathrm{gravel} \; \mathbf{U} \;\mathrm{goal}) \geq 0.9$, where $\mathrm{gravel}$ is an atomic proposition that is true for states having gravel as its feature, and $\mathrm{goal}$ is an atomic proposition that is true at each target state. Note that this side information does not necessarily enforce that the learner should reach the set of target states. Instead, if the learner reaches the target state, it should not drive on gravel with probability at least 

\paragraph{\textbf{Results}}
Figure~\ref{fig:traj_res} shows how the learner with side information avoids the gravel compared to the learner without side information.
Figure~\ref{fig:rew-evolution} further illustrates this result by empirically demonstrating that the proposed approach can efficiently take advantage of side information to compute policies that matches the expert's desired behavior.
Specifically, Figure~\ref{fig:rew-evolution} shows that the gain in the total reward of a learner without side information increases by $294\%$ with respect to a learner with side information. 
\myett{Additionally, it is important to note in Figure~\ref{fig:gridphoenix} how the initial state distribution of the demonstrator trajectories is different from the initial state distribution during the evaluation of the learned policies (Figure 7a). Nevertheless, despite these distinctions, the learned policies can effectively navigate toward points present in the expert demonstrations and then maximally mimic these trajectories.}




\section{Related work.} 
The closest work to ours is by~\cite{choi2011inverse}, where they extend classical maximum-margin-based IRL techniques for MDPs to POMDPs. However, even on MDPs, maximum-margin-based approaches cannot resolve the ambiguity caused by suboptimal demonstrations, and they work well when there is a single reward function that is clearly better than alternatives~\cite{osa2018algorithmic}. In contrast, we adopt causal entropy that has been shown~\cite{osa2018algorithmic, ziebart2010modeling} to alleviate these limitations on MDPs. Besides, ~\cite{choi2011inverse} rely on efficient off-the-shelf solvers to the forward problem. Instead, this paper also develops an algorithm that outperforms off-the-shelf solvers and can scale to POMDPs that are orders of magnitude larger compared to the examples in ~\cite{choi2011inverse}. Further, \cite{choi2011inverse} do not incorporate task specifications in their formulations.

One of the basic challenges in IRL, is that finding a reward function and a policy that induces a similar behavior to the expert is an ill-defined problem. 
Prior work has addressed this challenge using maximum margin formulations~\cite{ratliff2006maximum,abbeel2004apprenticeship,ng2000algorithms}, as well as probabilistic models to compute a likelihood of the expert demonstrations~\cite{ramachandran2007bayesian,ziebart2008maximum,ziebart2010modeling}.
We build on the latter approach and build on the maximum-causal-entropy IRL~\cite{zhou2017infinite,ziebart2010modeling,ziebart2013principle}, which brings algorithmic benefits to IRL in POMDPs as mentioned in the introduction.
We note that these maximum-causal-entropy IRL techniques assume that both the expert and the agent can fully observe the environment, and these approaches only apply for MDPs as opposed to POMDPs.

IRL under partial information has been studied in prior work~\cite{kitani2012activity,boularias2012structured,bogert2014multi,bogert2015toward,bogert2016expectation}.
Reference~\cite{boularias2012structured} considers the setting where the features of the reward function are partially specified as opposed to having partial information over the state of the environment.
The work in~\cite{kitani2012activity} considers a special case of POMDPs.
It only infers a distribution over the future trajectories of the expert given demonstrations as opposed to computing a policy that induces a similar behavior to the expert.
The works in~\cite{bogert2014multi,bogert2015toward,bogert2016expectation} assume that the states of the environment are either fully observable, or fully hidden to the learning agent. Therefore, these approaches also consider a special case of POMDPs, like in~\cite{kitani2012activity}.
We also note that none of these methods incorporate side information into IRL and do not provide guarantees on the performance of the policy with respect to a task specification.

The idea of using side information expressed in temporal logic to guide and augment IRL has been explored in some previous work. 
In~\cite{papusha2018inverse,wen2017learning}, the authors incorporate side information as in temporal logic specification to learn policies that induce a behavior similar to the expert demonstrations and satisfies the specification.
Reference~\cite{memarian2020active} iteratively infers an underlying task specification that is consistent with the expert demonstrations and learns a policy and a reward function that satisfies the task specification.
However, these methods also assume full information for both the expert and the agent.

\section{Conclusion}

We develop an algorithm for inverse reinforcement learning under partial observation.
We empirically demonstrate that by incorporating task specifications into the learning process, we can alleviate the information asymmetry between the expert and the learner while increasing the data efficiency of the learning scheme. Further, we empirically demonstrate that our main routine $\texttt{SCPForward}$, used inside the IRL algorithm, solves the forward problem in a scalable manner and outperforms state-of-the-art POMDP solvers on instances with a large number of states, observations, and transitions.

\paragraph{\textbf{Work Limitations}} This work assumes that the transition and observation functions of the POMDP are known to the algorithm. Future work will investigate removing this assumption and developing model-free-based approaches. We will also integrate the framework with more expressive neural-network-based reward functions.

\paragraph{\textbf{Acknowledgements.}}
Research was sponsored by the Army Research Laboratory and Office of Naval Research accomplished under cooperative agreement number(s) ARL W911NF-20-2-0132, ARL W911NF-19-2-0285 and ONR N00014-22-1-2254. The views and conclusions contained in this document are those of the authors and should not be interpreted as representing the official policies; either expressed or implied, of the Army Research Laboratory, Office of Naval Research, or the U.S. Government. The U.S. Government is authorized to reproduce and distribute reprints for Government purposed notwithstanding any copyright notation herein.

\bibliographystyle{elsarticle-num}
\bibliography{bibliography.bib}

\begin{appendices}\label{appendix}

In this appendix, we provide supplementary derivations for the results in the paper and more details on the numerical experiments.

\section{Concavity of Causal Entropy and Derivations of the Bellman Constraints}

In this section, we first recall the obtained expression of the causal entropy $H^\gamma_\policy$ as a function of the visitation counts $\StateOccup^\discount_\policy$ and $\StateActionOccup^\discount_\policy$.
We then prove the concavity of the causal entropy, which enables convex-optimization-based formulation of the task-guided inverse reinforcement learning (IRL) problem. 
Then, we provide additional details on the derivation of the affine constraint implied by the \emph{Bellman flow constraint}.

\paragraph{\textbf{Concave Causal Entropy}}
We first recall the definitions of the state and state-action visitation counts.
For a policy $\policy$, state $\state$, and action $\action$, the discounted state visitation counts are defined by $\StateOccup^\discount_\policy (\state) \triangleq \mathbb{E}_{S_t} [\sum_{t=1}^\infty \discount^t \mathbbm{1}_{\{S_t = \state\}}]$ and the discounted state-action visitation counts are defined by $\StateActionOccup^\discount_\policy (\state,\action) \triangleq \mathbb{E}_{A_t, S_t} [\sum_{t=1}^\infty \discount^t \mathbbm{1}_{\{S_t = \state, A_t = \action\}}],$ where $\mathbbm{1}_{\{\cdot\}}$ is the indicator function and $t$ is the time step.
From the definitions of the state and state-action visitation counts $\StateOccup^\discount_\policy$ and $\StateActionOccup^\discount_\policy$, it is straightforward to deduce that $\StateActionOccup^\discount_\policy(\state,\action) = \policy_{\state,\action}\StateOccup^\discount_\policy(\state)$, where $\policy_{\state,\action} = \mathbb{P}[A_t=a | S_t = s] $. 
We use the visitation counts to prove in Section~$4$ that
\begin{align}
    H_\policy^\discount = \sum_{(\state,\action)\in \States \times \Actions}  -(\log \pi_{\state,\action}) \pi_{\state,\action} \StateOccup^\discount_\policy (\state) 
&= \sum_{(\state,\action)\in \States \times \Actions} -\log \frac{\StateActionOccup^\discount_\policy(\state,\action)}{\StateOccup^\discount_\policy(\state)} \StateActionOccup^\discount_\policy(\state,\action), \nonumber
\end{align}%
where the last inequality is obtained by using that $\pi_{\state,\action} = \StateActionOccup^\discount_\policy(\state,\action) /  \StateOccup^\discount_\policy(\state)$. We claim that $H^\discount_\policy$ is a \emph{concave} fucntion of the visitation counts. Thus, we want to show that the function $f(\StateActionOccup^\gamma_\policy, \StateOccup^\gamma_\policy) = \sum_{(\state,\action)\in \States \times \Actions} -\log \frac{\StateActionOccup^\discount_\policy(\state,\action)}{\StateOccup^\discount_\policy(\state)} \StateActionOccup^\discount_\policy(\state,\action)$ is concave. To this end, consider any $\lambda \in (0,1)$ and the two sets of variables $\StateActionOccup^\gamma_\policy, \StateOccup^\gamma_\policy$ and $\bar{\StateActionOccup}^\gamma_\policy, \bar{\StateOccup}^\gamma_\policy$. 
Then, we have the following result:
\begin{align}
    &f(\lambda \StateActionOccup^\discount_\policy + (1-\lambda)  \bar{\StateActionOccup}^\discount_\policy, \lambda \bar{\StateOccup}^\discount_\policy + (1-\lambda)  \bar{\StateOccup}^\discount_\policy) \nonumber\\
    &\quad= \sum_{(\state,\action)\in \States \times \Actions} -\log \frac{\lambda \StateActionOccup^\discount_\policy(\state,\action) + (1-\lambda) \bar{\StateActionOccup}^\discount_\policy(\state,\action)}{\lambda \StateOccup^\discount_\policy(\state) + (1-\lambda) \bar{\StateOccup}^\discount_\policy(\state,\action)} (\lambda \StateActionOccup^\discount_\policy(\state,\action) + (1-\lambda) \bar{\StateActionOccup}^\discount_\policy(\state,\action)) \nonumber \\
    &\quad \geq \sum_{(\state,\action)\in \States \times \Actions} -\lambda \StateActionOccup^\discount_\policy(\state,\action) \log \frac{\lambda \StateActionOccup^\discount_\policy(\state,\action)}{\lambda \StateOccup^\discount_\policy(\state,\action)} - (1-\lambda) \bar{\StateActionOccup}^\discount_\policy(\state,\action) \log \frac{(1-\lambda) \bar{\StateActionOccup}^\discount_\policy(\state,\action)}{(1-\lambda) \bar{\StateOccup}^\discount_\policy(\state,\action)} \nonumber\\
    &\quad = \sum_{(\state,\action)\in \States \times \Actions} -\lambda \StateActionOccup^\discount_\policy(\state,\action) \log \frac{ \StateActionOccup^\discount_\policy(\state,\action)}{ \StateOccup^\discount_\policy(\state,\action)} - (1-\lambda) \bar{\StateActionOccup}^\discount_\policy(\state,\action) \log \frac{ \bar{\StateActionOccup}^\discount_\policy(\state,\action)}{ \bar{\StateOccup}^\discount_\policy(\state,\action)} \nonumber\\
    &\quad = \lambda f(\StateActionOccup^\discount_\policy, \StateOccup^\discount_\policy) + (1-\lambda) f(\bar{\StateActionOccup}^\discount_\policy,\bar{\StateOccup}^\discount_\policy), \nonumber
\end{align}
where the first inequality is obtained by applying to the well-known \emph{log-sum inequality}, i.e., $$x_1 \log \frac{x_1}{y_1} + x_2 \log \frac{x_2}{y_2} \geq (x_1+x_2) \log \frac{x_1+x_2}{y_1+y_2},$$ for nonnegative numbers $x_1,x_2, y_1, y_2$. 
Specifically, we apply the substitution $x_1 =  \lambda \StateActionOccup^\discount_\policy$, $y_1 = \lambda \StateOccup^\discount_\policy$, $x_2 = (1-\lambda) \bar{\StateActionOccup}^\discount_\policy$, and $y_2 = (1-\lambda) \bar{\StateOccup}^\discount_\policy$.
Note that the inequality 
$$f(\lambda \StateActionOccup^\discount_\policy + (1-\lambda)  \bar{\StateActionOccup}^\discount_\policy, \lambda \bar{\StateOccup}^\discount_\policy + (1-\lambda)  \bar{\StateOccup}^\discount_\policy) \geq \lambda f(\StateActionOccup^\discount_\policy, \StateOccup^\discount_\policy) + (1-\lambda) f(\bar{\StateActionOccup}^\discount_\policy,\bar{\StateOccup}^\discount_\policy)$$
implies that $f(\StateActionOccup^\gamma_\policy, \StateOccup^\gamma_\policy)$ is concave in $\StateActionOccup^\gamma_\policy,$ and  $\StateOccup^\gamma_\policy$.

\paragraph{\textbf{Bellman Flow Constraint}} For the visitation count variables to correspond to a valid policy generating actions in the POMDP $\Pomdp$ , $\StateActionOccup^\discount_\policy$ and $\StateOccup^\discount_\policy$ must satisfy the bellman flow constraint given by
\begin{align}
    \StateOccup^\discount_{\policy} (\state)  &=
    \mathbb{E}_{S^\policy_t} \Big[ \sum_{t=0}^\infty \discount^t \mathbbm{1}_{\{S^\policy_t = \state\} }\Big] \nonumber \\
    &= \Initdist(\state) + \discount \mathbb{E}_{S^\policy_t} \Big[ \sum_{t=0}^\infty \discount^{t} \mathbbm{1}_{\{S^\policy_{t+1} = \state\} }\Big] \nonumber \\
    &= \Initdist(\state) + \discount  \sum_{t=0}^\infty \sum_{\state' \in \States} \sum_{\action \in \Actions} \discount^t \Transition(\state | \state',\action) \mathbb{P}[S^\policy_t=\state', A^\policy_t = \action] \nonumber\\
    &= \Initdist(\state) + \discount \sum_{\state' \in \States} \sum_{\action \in \Actions} \Transition(\state | \state',\action) \StateActionOccup^\discount_{\policy} (\state',\action).\nonumber
\end{align}%

\section{Experimental Tasks}

In this section, we first provide a detailed description of the POMDP models used in the benchmark. 
The simulations on the benchmark examples empirically demonstrate that side information alleviates the information asymmetry, and more memory leads to more performant policies. 
Then, we provide additional numerical simulations supporting the claim that $\texttt{SCPForward}$ is sound and yields better scalability than off-the-shelf solvers for the \emph{forward problem}, i.e., computing an optimal policy on a POMDP for a given reward function.

\subsection{Computation Resources and External Assets}
All the experiments of this paper were performed on a computer with an Intel Core $i9$-$9900$ CPU $3.1$GHz $\times 16$ processors and $31.2$ Gb of RAM. 
All the implementations are written and tested in Python $3.8$, and we attach the code with the supplementary material. 

\paragraph{Required Tools. } Our implementation requires \emph{Stormpy} of \emph{Storm}~\cite{hensel2020probabilistic} and \emph{Gurobipy} of \emph{Gurobi} $9.1$~\cite{gurobi}. 
On one hand, we use \emph{Storm}, a tool for model checking, to parse POMDP file specifications, to compute the product POMDP with the finite state controller in order  to reduce the synthesis problem to the synthesis of memoryless policies, and to compute the set $\target$ of target states satisfying a specification $\varphi$ via graph preprocessing. 
On the other hand, we use \emph{Gurobi} to solve both the linearized problem in  ($7$) and the feasible solution of the Bellman flow constraint needed for the \emph{verification step}.

\paragraph{Off-The-Shelf Solvers for Forward Problem. } In order to show the scalability of the developed algorithm $\texttt{SCPForward}$, we compare it to state-of-the-art POMDP solvers $\texttt{SolvePOMDP}$~\cite{walraven2017accelerated}, $\texttt{SARSOP}$~\cite{kurniawati2008sarsop}, and  $\texttt{PRISM-POMDP}$~\cite{norman2017verification}. 
The solver $\texttt{SolvePOMDP}$ implements both exact and approximate value iterations via incremental pruning~\cite{Cassandra97incrementalpruning} combined with state-of-the-art vector pruning methods~\cite{walraven2017accelerated}. 
Finally, $\texttt{PrismPOMDP}$ discretizes the belief state and adopts a finite memory strategy to find an approximate solution of the forward problem. For all the solvers above, we use the default settings except from the timeout enforced to be $3600$ seconds. 
These solvers are not provided with our implementation. However, we provide the POMDP models that each of the solvers can straightforwardly use. Further details are provided in the readme files of our implementation.

\subsection{Benchmark Set}
We evaluate the proposed learning algorithm on several POMDP instances adapted from~\cite{junges2020enforcing}. 
We attached the modified instances in our code with the automatically generated models for each off-the-shelf solver that the reader can straightforwardly use to reproduce Table~\ref{tab:ToolComp}. 
The reader can refer to Table~\ref{tab:ToolComp} for the number of states, observations, and transitions of each environment of the benchmark set. 
In all the examples, we gather $10$ trajectories from an expert that can fully observe its current state in the environment and an expert having partial observation of the environment. 
Our algorithm learns reward functions from these trajectories under different memory policies and high-level side information.

\clearpage
\paragraph{\textbf{Rocks Instance}} 
In the environment $\mathrm{Rocks}$, an agent navigates in a gridworld to sample at least one valuable rock (if a valuable rock is in the grid)  over the two possibly dangerous rocks, without any failures. 
When at least one valuable rock has been collected, or the agent realizes that all the rocks are dangerous, it needs to get to an exit point to terminate the mission. 
The partial observability is due to the agent can only observe if its current location is an exit point or a dangerous rock. Furthermore, the agent has noisy sensors enabling sampling neighbor cells. 

We consider three feature functions. 
The first feature provides a positive reward when reaching the exit point with at least one valuable rock or no rocks when all of them are dangerous. 
The second feature provides a negative reward when the agent is at the location of a dangerous rock. 
Finally, the third feature penalizes each action taken with a negative reward to promote reaching the exit point as soon as possible.

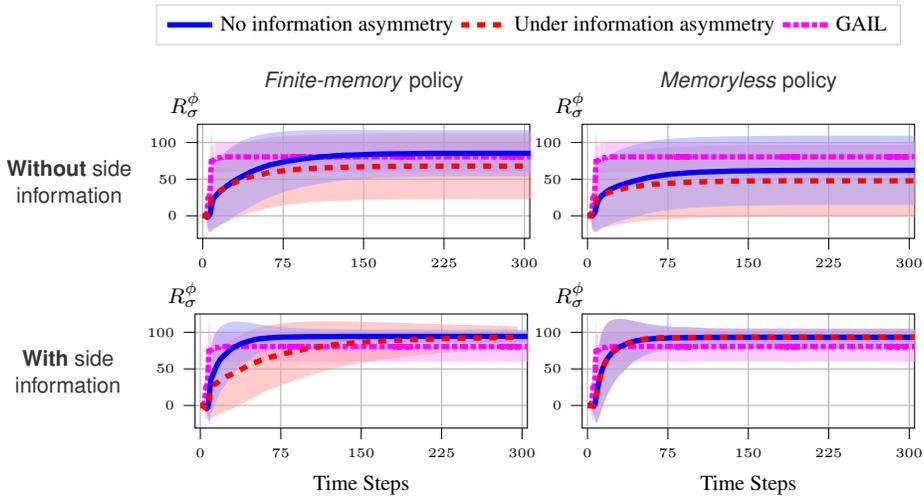
\begin{figure}[!hbt]
\centering
    \begin{subfigure}[t]{\textwidth}
        \centering\hspace*{1.8cm}\begin{tikzpicture}
\definecolor{color1}{HTML}{FF0000}
\definecolor{color0}{HTML}{0000FF}
\definecolor{color2}{HTML}{FF00FF}
\begin{customlegend}[legend columns=3,legend style={
  fill opacity=0.8,
  draw opacity=1,
  text opacity=1,
  at={(1.3,0.4)},
  anchor=south,
  draw=white!80!black,
  scale=0.40,
    font=\titlesize,
  mark options={scale=0.5},
},legend entries={No information asymmetry, Under information asymmetry, GAIL}]

\addlegendimage{line width=2pt, color0}
\addlegendimage{line width=2pt, color1, dashed}
\addlegendimage{line width=2pt, color2, densely dashdotted}
\end{customlegend}
\end{tikzpicture}
    \end{subfigure}\hfill%
    \vspace*{0.20cm}
    \hspace*{-2mm}
    \centering
    \begin{subfigure}[t]{0.541\textwidth}
        \centering
        \centering\captionsetup{width=.95\linewidth}%
\begin{tikzpicture}

\definecolor{color1}{HTML}{FF0000}
\definecolor{color0}{HTML}{0000FF}
\definecolor{color2}{HTML}{FF00FF}

\begin{axis}[
legend cell align={left},
legend columns=2,
legend style={
  fill opacity=0.8,
  draw opacity=1,
  text opacity=1,
  at={(1.15,1.25)},
  anchor=south,
    ticklabel style = {font=\tiny},
  draw=white!80!black,
  font=\fontsize{7.4}{7.4}\selectfont,
  scale=0.40
},
title={$R_\policy^\phi$},
title style={yshift=-1.5ex,xshift=-15ex,font=\titlesize},
width=6.cm,
height=3.1cm,
tick align=outside,
tick pos=left,
no markers,
every axis plot/.append style={line width=2pt},
x grid style={white!69.0196078431373!black},
xlabel={\footnotesize{Time steps}},
xlabel={\textcolor{gray!40!black}{\arial \textit{Finite-memory} policy}},
xlabel style={font=\titlesize,yshift=18ex},
xmajorgrids,
xmin=-4.95, xmax=305,
xtick style={color=black},
y grid style={white!69.0196078431373!black},
ylabel={\footnotesize{Mean  accumulated  reward}},
ylabel={\textcolor{gray!40!black}{\arial\textbf{Without} side} \\\textcolor{gray!40!black}{\arial information}},
ylabel style={rotate=-90,align=center,font=\titlesize},
xtick={0,75,150,225,300},
ymajorgrids,
ymin=-30.8979836098955, ymax=125,
ytick style={color=black}
]

\addplot[color2,densely dashdotted] table[x=x,y=y,mark=none] {rock_mdp_fwd_gail.tex};
\addplot [name path=upper2,draw=none] table[x=x,y expr=\thisrow{y}+\thisrow{err}] {rock_mdp_fwd_gail.tex};
\addplot [name path=lower2,draw=none] table[x=x,y expr=\thisrow{y}-\thisrow{err}] {rock_mdp_fwd_gail.tex};
\addplot [fill=color2!40,opacity=0.5] fill between[of=upper2 and lower2];

\addplot[color0] table[x=x,y=y,mark=none] {rock_mem10_trajsize10pomdp_irl.tex};
\addplot[color1,dashed] table[x=x,y=y,mark=none] {rock_mem10_trajsize10mdp_irl.tex};

\addplot [name path=upper1,draw=none] table[x=x,y expr=\thisrow{y}+\thisrow{err}]{rock_mem10_trajsize10mdp_irl.tex};
\addplot [name path=lower1,draw=none] table[x=x,y expr=\thisrow{y}-\thisrow{err}] {rock_mem10_trajsize10mdp_irl.tex};
\addplot [fill=color1!40,opacity=0.5] fill between[of=upper1 and lower1];

\addplot [name path=upper,draw=none] table[x=x,y expr=\thisrow{y}+\thisrow{err}] {rock_mem10_trajsize10pomdp_irl.tex};
\addplot [name path=lower,draw=none] table[x=x,y expr=\thisrow{y}-\thisrow{err}] {rock_mem10_trajsize10pomdp_irl.tex};
\addplot [fill=color0!40,opacity=0.5] fill between[of=upper and lower];

\end{axis}

\end{tikzpicture}
    \end{subfigure}%
    \begin{subfigure}[t]{0.511\textwidth}
        \centering
        \centering\captionsetup{width=.95\linewidth}%
\begin{tikzpicture}

\definecolor{color1}{HTML}{FF0000}
\definecolor{color0}{HTML}{0000FF}
\definecolor{color2}{HTML}{FF00FF}
\begin{axis}[
legend cell align={left},
legend columns=2,
legend style={
  fill opacity=0.8,
  draw opacity=1,
  text opacity=1,
  at={(0.5,1)},
  anchor=south,
  ticklabel style = {font=\tiny},
  draw=white!80!black,
  font=\fontsize{6}{6}\selectfont,
  scale=0.40
},
title={$R_\policy^\phi$},
title style={yshift=-1.5ex,xshift=-15ex,font=\titlesize},
xshift=0.1cm,
width=6.cm,
height=3.1cm,
tick align=outside,
tick pos=left,
no markers,
every axis plot/.append style={line width=2pt},
x grid style={white!69.0196078431373!black},
xlabel={\footnotesize{Time steps}},
xlabel={},
xlabel={\textcolor{gray!40!black}{\arial \textit{Memoryless} policy}},
xlabel style={font=\titlesize,yshift=18ex},
xtick={0,75,150,225,300},
xmajorgrids,
xmin=-4.95, xmax=305,
xtick style={color=black},
y grid style={white!69.0196078431373!black},
ylabel={\footnotesize{Mean  accumulated  reward}},
ylabel={},
ymajorgrids,
ymin=-30.8979836098955, ymax=125,
ytick style={color=black}
]

\addplot[color2,densely dashdotted] table[x=x,y=y,mark=none] {rock_mdp_fwd_gail.tex};
\addplot [name path=upper2,draw=none] table[x=x,y expr=\thisrow{y}+\thisrow{err}] {rock_mdp_fwd_gail.tex};
\addplot [name path=lower2,draw=none] table[x=x,y expr=\thisrow{y}-\thisrow{err}] {rock_mdp_fwd_gail.tex};
\addplot [fill=color2!40,opacity=0.5] fill between[of=upper2 and lower2];

\addplot[color0] table[x=x,y=y,mark=none] {rock_mem1_trajsize10pomdp_irl.tex};
\addplot[color1,dashed] table[x=x,y=y,mark=none] {rock_mem1_trajsize10mdp_irl.tex};

\addplot [name path=upper1,draw=none] table[x=x,y expr=\thisrow{y}+\thisrow{err}]{rock_mem1_trajsize10mdp_irl.tex};
\addplot [name path=lower1,draw=none] table[x=x,y expr=\thisrow{y}-\thisrow{err}] {rock_mem1_trajsize10mdp_irl.tex};
\addplot [fill=color1!40,opacity=0.5] fill between[of=upper1 and lower1];

\addplot [name path=upper,draw=none] table[x=x,y expr=\thisrow{y}+\thisrow{err}] {rock_mem1_trajsize10pomdp_irl.tex};
\addplot [name path=lower,draw=none] table[x=x,y expr=\thisrow{y}-\thisrow{err}] {rock_mem1_trajsize10pomdp_irl.tex};
\addplot [fill=color0!40,opacity=0.5] fill between[of=upper and lower];

\end{axis}

\end{tikzpicture}
    \end{subfigure}%
    \vfill
    \hspace*{-2mm}
    \centering
        \begin{subfigure}[t]{0.541\textwidth}
        \centering
        \centering\captionsetup{width=.95\linewidth}%
\begin{tikzpicture}

\definecolor{color1}{HTML}{FF0000}
\definecolor{color0}{HTML}{0000FF}
\definecolor{color2}{HTML}{FF00FF}

\begin{axis}[
legend cell align={left},
legend columns=2,
legend style={
  fill opacity=0.8,
  draw opacity=1,
  text opacity=1,
  at={(0.5,1)},
  anchor=south,
  ticklabel style = {font=\tiny},
  draw=white!80!black,
  font=\fontsize{6}{6}\selectfont,
  scale=0.40
},
title={\textbf{Finite-memory} policy \emph{with} side information},
title={$R_\policy^\phi$},
title style={yshift=-1.5ex,xshift=-15ex,font=\titlesize},
yshift=0.5cm,
width=6.cm,
height=3.1cm,
tick align=outside,
tick pos=left,
no markers,
every axis plot/.append style={line width=2pt},
x grid style={white!69.0196078431373!black},
xlabel={\titlesize{Time Steps}},
xmajorgrids,
xmin=-4.95, xmax=305,
xtick style={color=black},
y grid style={white!69.0196078431373!black},
xtick={0,75,150,225,300},
ylabel={\footnotesize{Mean accumulated  reward}},
ylabel={\titlesize{$R_\policy^\phi$}},
ylabel={\textcolor{gray!40!black}{\;\;\,\arial\textbf{With} side\;\;\,} \\\textcolor{gray!40!black}{\arial information}},
ylabel style={rotate=-90,align=center,font=\titlesize},
ymajorgrids,
ymin=-30.8979836098955, ymax=125,
ytick style={color=black}
]

\addplot[color2,densely dashdotted] table[x=x,y=y,mark=none] {rock_mdp_fwd_gail.tex};
\addplot [name path=upper2,draw=none] table[x=x,y expr=\thisrow{y}+\thisrow{err}] {rock_mdp_fwd_gail.tex};
\addplot [name path=lower2,draw=none] table[x=x,y expr=\thisrow{y}-\thisrow{err}] {rock_mdp_fwd_gail.tex};
\addplot [fill=color2!40,opacity=0.5] fill between[of=upper2 and lower2];

\addplot[color0] table[x=x,y=y,mark=none] {rock_mem10_trajsize10pomdp_irl_si.tex};
\addplot[color1,dashed] table[x=x,y=y,mark=none] {rock_mem10_trajsize10mdp_irl_si.tex};

\addplot [name path=upper1,draw=none] table[x=x,y expr=\thisrow{y}+\thisrow{err}]{rock_mem10_trajsize10mdp_irl_si.tex};
\addplot [name path=lower1,draw=none] table[x=x,y expr=\thisrow{y}-\thisrow{err}] {rock_mem10_trajsize10mdp_irl_si.tex};
\addplot [fill=color1!40,opacity=0.5] fill between[of=upper1 and lower1];

\addplot [name path=upper,draw=none] table[x=x,y expr=\thisrow{y}+\thisrow{err}] {rock_mem10_trajsize10pomdp_irl_si.tex};
\addplot [name path=lower,draw=none] table[x=x,y expr=\thisrow{y}-\thisrow{err}] {rock_mem10_trajsize10pomdp_irl_si.tex};
\addplot [fill=color0!40,opacity=0.5] fill between[of=upper and lower];

\end{axis}

\end{tikzpicture}
    \end{subfigure}%
    \begin{subfigure}[t]{0.511\textwidth}
        \centering
        \centering\captionsetup{width=.95\linewidth}%
\begin{tikzpicture}

\definecolor{color1}{HTML}{FF0000}
\definecolor{color0}{HTML}{0000FF}
\definecolor{color2}{HTML}{FF00FF}
\begin{axis}[
legend cell align={left},
legend columns=2,
legend style={
  fill opacity=0.8,
  draw opacity=1,
  text opacity=1,
  at={(0.5,1)},
  anchor=south,
  ticklabel style = {font=\tiny},
  draw=white!80!black,
  font=\fontsize{6}{6}\selectfont,
  scale=0.40
},
title={$R_\policy^\phi$},
title style={yshift=-1.5ex,xshift=-15ex,font=\titlesize},
xshift=0.1cm,
yshift = 0.5cm,
width=6.cm,
height=3.1cm,
tick align=outside,
tick pos=left,
no markers,
every axis plot/.append style={line width=2pt},
x grid style={white!69.0196078431373!black},
xlabel={\titlesize{Time Steps}},
xmajorgrids,
xmin=-4.95, xmax=305,
xtick style={color=black},
y grid style={white!69.0196078431373!black},
ylabel={\footnotesize{Mean  accumulated  reward}},
ylabel={},
ymajorgrids,
xtick={0,75,150,225,300},
ymin=-30.8979836098955, ymax=125,
ytick style={color=black}
]

\addplot[color2,densely dashdotted] table[x=x,y=y,mark=none] {rock_mdp_fwd_gail.tex};
\addplot [name path=upper2,draw=none] table[x=x,y expr=\thisrow{y}+\thisrow{err}] {rock_mdp_fwd_gail.tex};
\addplot [name path=lower2,draw=none] table[x=x,y expr=\thisrow{y}-\thisrow{err}] {rock_mdp_fwd_gail.tex};
\addplot [fill=color2!40,opacity=0.5] fill between[of=upper2 and lower2];

\addplot[color0] table[x=x,y=y,mark=none] {rock_mem1_trajsize10pomdp_irl_si.tex};
\addplot[color1,dashed] table[x=x,y=y,mark=none] {rock_mem1_trajsize10mdp_irl_si.tex};

\addplot [name path=upper1,draw=none] table[x=x,y expr=\thisrow{y}+\thisrow{err}]{rock_mem1_trajsize10mdp_irl_si.tex};
\addplot [name path=lower1,draw=none] table[x=x,y expr=\thisrow{y}-\thisrow{err}] {rock_mem1_trajsize10mdp_irl_si.tex};
\addplot [fill=color1!40,opacity=0.5] fill between[of=upper1 and lower1];

\addplot [name path=upper,draw=none] table[x=x,y expr=\thisrow{y}+\thisrow{err}] {rock_mem1_trajsize10pomdp_irl_si.tex};
\addplot [name path=lower,draw=none] table[x=x,y expr=\thisrow{y}-\thisrow{err}] {rock_mem1_trajsize10pomdp_irl_si.tex};
\addplot [fill=color0!40,opacity=0.5] fill between[of=upper and lower];
\end{axis}

\end{tikzpicture}
    \end{subfigure}%
    \vspace*{-0.47cm}
    \caption{Representative results on the $\mathrm{Rock}$ example showing the reward of the policies under the true reward function ($R_\policy^\phi$) versus the time steps.}
    \label{fig:rock}
\end{figure}%

We compare scenarios with no side information and the a priori knowledge on the task such as \emph{the agent eventually reaches an exit point with a probability greater than $0.995$.} Figure~\ref{fig:rock} supports our claim that side information indeed alleviates the information asymmetry between the expert and the agent.
Additionally, we also observe that incorporating memory leads to more performant policies in terms of the mean accumulated reward.

\clearpage
\paragraph{Obstacle Instance. } 
In the environment $\mathrm{Obstacle}[n]$, an agent must find an exit in a gridworld without colliding with any of the five static obstacles in the grid. 
The agent only observes whether the current position is an obstacle or an exit state. The parameter $n$ specifies the dimension of the grid.

Similar to the $\mathrm{Rocks}$ example, the agent receives a positive reward if it successfully exits the gridworld and a negative reward for every taken action or colliding with an obstacle.

As for the side information, we specify in temporal logic that while learning the reward, \emph{the agent should not collide any obstacles until it reaches an exit point with a probability greater than $0.9$.}

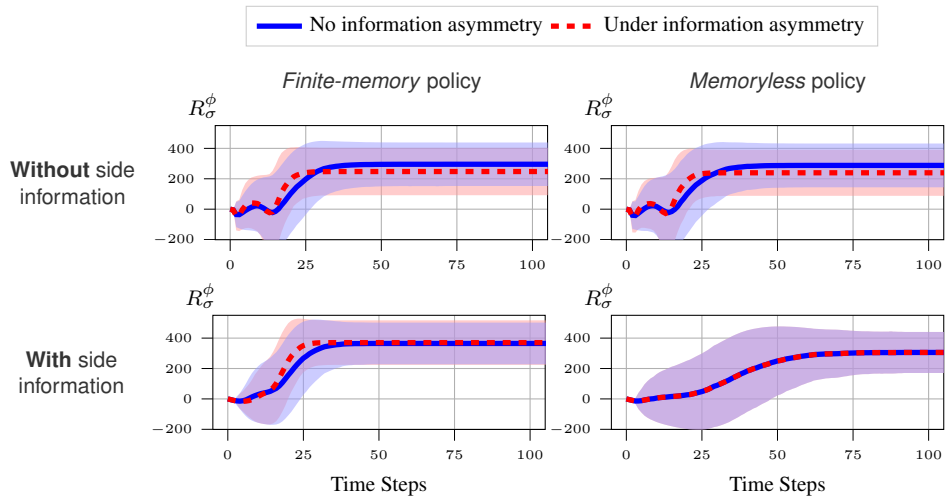
\begin{figure}[!hbt]
\centering
    \begin{subfigure}[t]{\textwidth}
        \centering\hspace*{2.4cm}\begin{tikzpicture}
\definecolor{color1}{HTML}{FF0000}
\definecolor{color0}{HTML}{0000FF}
\definecolor{color2}{HTML}{FF00FF}
\begin{customlegend}[legend columns=3,legend style={
  fill opacity=0.8,
  draw opacity=1,
  text opacity=1,
  at={(1.3,0.4)},
  anchor=south,
  draw=white!80!black,
  scale=0.40,
    font=\titlesize,
  mark options={scale=0.5},
},legend entries={No information asymmetry, Under information asymmetry}]

\addlegendimage{line width=2pt, color0}
\addlegendimage{line width=2pt, color1, dashed}
\addlegendimage{line width=2pt, color2, densely dashdotted}
\end{customlegend}
\end{tikzpicture}
    \end{subfigure}\hfill%
    \vspace*{0.20cm}
    \hspace*{-3mm}
    \centering
    \begin{subfigure}[t]{0.541\textwidth}
        \centering
        \centering\captionsetup{width=.95\linewidth}%
\begin{tikzpicture}

\definecolor{color1}{HTML}{FF0000}
\definecolor{color0}{HTML}{0000FF}

\begin{axis}[
legend cell align={left},
legend columns=2,
legend style={
  fill opacity=0.8,
  draw opacity=1,
  text opacity=1,
  at={(1.15,1.25)},
  anchor=south,
    ticklabel style = {font=\tiny},
  draw=white!80!black,
  font=\fontsize{7.4}{7.4}\selectfont,
  scale=0.40
},
title={$R_\policy^\phi$},
title style={yshift=-1.5ex,xshift=-15ex,font=\titlesize},
width=6.cm,
height=3.1cm,
tick align=outside,
tick pos=left,
no markers,
every axis plot/.append style={line width=2pt},
x grid style={white!69.0196078431373!black},
xlabel={\footnotesize{Time steps}},
xlabel={\textcolor{gray!40!black}{\arial \textit{Finite-memory} policy}},
xlabel style={font=\titlesize,yshift=18ex},
xmajorgrids,
xmin=-4.95, xmax=105,
xtick style={color=black},
y grid style={white!69.0196078431373!black},
ylabel={\footnotesize{Mean  accumulated  reward}},
ylabel={\textcolor{gray!40!black}{\arial\textbf{Without} side} \\\textcolor{gray!40!black}{\arial information}},
ylabel style={rotate=-90,align=center,font=\titlesize},
xtick={0,25,50,75,100},
ymajorgrids,
ymin=-200.8979836098955, ymax=550,
ytick style={color=black}
]

\addplot[color0] table[x=x,y=y,mark=none] {obstacle_mem5_trajsize10pomdp_irl.tex};
\addplot[color1,dashed] table[x=x,y=y,mark=none] {obstacle_mem5_trajsize10mdp_irl.tex};

\addplot [name path=upper1,draw=none] table[x=x,y expr=\thisrow{y}+\thisrow{err}]{obstacle_mem5_trajsize10mdp_irl.tex};
\addplot [name path=lower1,draw=none] table[x=x,y expr=\thisrow{y}-\thisrow{err}] {obstacle_mem5_trajsize10mdp_irl.tex};
\addplot [fill=color1!40,opacity=0.5] fill between[of=upper1 and lower1];

\addplot [name path=upper,draw=none] table[x=x,y expr=\thisrow{y}+\thisrow{err}] {obstacle_mem5_trajsize10pomdp_irl.tex};
\addplot [name path=lower,draw=none] table[x=x,y expr=\thisrow{y}-\thisrow{err}] {obstacle_mem5_trajsize10pomdp_irl.tex};
\addplot [fill=color0!40,opacity=0.5] fill between[of=upper and lower];

\end{axis}

\end{tikzpicture}
    \end{subfigure}%
    \begin{subfigure}[t]{0.541\textwidth}
        \centering
        \centering\captionsetup{width=.95\linewidth}%
\begin{tikzpicture}

\definecolor{color1}{HTML}{FF0000}
\definecolor{color0}{HTML}{0000FF}

\begin{axis}[
legend cell align={left},
legend columns=2,
legend style={
  fill opacity=0.8,
  draw opacity=1,
  text opacity=1,
  at={(0.5,1)},
  anchor=south,
  ticklabel style = {font=\tiny},
  draw=white!80!black,
  font=\fontsize{6}{6}\selectfont,
  scale=0.40
},
title={$R_\policy^\phi$},
title style={yshift=-1.5ex,xshift=-15ex,font=\titlesize},
xshift=0.1cm,
width=6.cm,
height=3.1cm,
tick align=outside,
tick pos=left,
no markers,
every axis plot/.append style={line width=2pt},
x grid style={white!69.0196078431373!black},
xlabel={\footnotesize{Time steps}},
xlabel={},
xlabel={\textcolor{gray!40!black}{\arial \textit{Memoryless} policy}},
xlabel style={font=\titlesize,yshift=18ex},
xtick={0,25,50,75,100},
xmajorgrids,
xmin=-4.95, xmax=105,
xtick style={color=black},
y grid style={white!69.0196078431373!black},
ylabel={\footnotesize{Mean  accumulated  reward}},
ylabel={},
ymajorgrids,
ymin=-200.8979836098955, ymax=550,
ytick style={color=black}
]

\addplot[color0] table[x=x,y=y,mark=none] {obstacle_mem1_trajsize10pomdp_irl.tex};
\addplot[color1,dashed] table[x=x,y=y,mark=none] {obstacle_mem1_trajsize10mdp_irl.tex};

\addplot [name path=upper1,draw=none] table[x=x,y expr=\thisrow{y}+\thisrow{err}]{obstacle_mem1_trajsize10mdp_irl.tex};
\addplot [name path=lower1,draw=none] table[x=x,y expr=\thisrow{y}-\thisrow{err}] {obstacle_mem1_trajsize10mdp_irl.tex};
\addplot [fill=color1!40,opacity=0.5] fill between[of=upper1 and lower1];

\addplot [name path=upper,draw=none] table[x=x,y expr=\thisrow{y}+\thisrow{err}] {obstacle_mem1_trajsize10pomdp_irl.tex};
\addplot [name path=lower,draw=none] table[x=x,y expr=\thisrow{y}-\thisrow{err}] {obstacle_mem1_trajsize10pomdp_irl.tex};
\addplot [fill=color0!40,opacity=0.5] fill between[of=upper and lower];

\end{axis}

\end{tikzpicture}
    \end{subfigure}%
    \vfill
    \hspace*{-3mm}
    \centering
        \begin{subfigure}[t]{0.541\textwidth}
        \centering
        \centering\captionsetup{width=.95\linewidth}%
\begin{tikzpicture}

\definecolor{color1}{HTML}{FF0000}
\definecolor{color0}{HTML}{0000FF}

\begin{axis}[
legend cell align={left},
legend columns=2,
legend style={
  fill opacity=0.8,
  draw opacity=1,
  text opacity=1,
  at={(0.5,1)},
  anchor=south,
  ticklabel style = {font=\tiny},
  draw=white!80!black,
  font=\fontsize{6}{6}\selectfont,
  scale=0.40
},
title={\textbf{Finite-memory} policy \emph{with} side information},
title={$R_\policy^\phi$},
title style={yshift=-1.5ex,xshift=-15ex,font=\titlesize},
yshift=0.5cm,
width=6.cm,
height=3.1cm,
tick align=outside,
tick pos=left,
no markers,
every axis plot/.append style={line width=2pt},
x grid style={white!69.0196078431373!black},
xlabel={\titlesize{Time Steps}},
xmajorgrids,
xmin=-4.95, xmax=105,
xtick style={color=black},
y grid style={white!69.0196078431373!black},
xtick={0,25,50,75,100},
ylabel={\footnotesize{Mean accumulated  reward}},
ylabel={\titlesize{$R_\policy^\phi$}},
ylabel={\textcolor{gray!40!black}{\;\;\,\arial\textbf{With} side\;\;\,} \\\textcolor{gray!40!black}{\arial information}},
ylabel style={rotate=-90,align=center,font=\titlesize},
ymajorgrids,
ymin=-200.8979836098955, ymax=550,
ytick style={color=black}
]

\addplot[color0] table[x=x,y=y,mark=none] {obstacle_mem5_trajsize10pomdp_irl_si.tex};
\addplot[color1,dashed] table[x=x,y=y,mark=none] {obstacle_mem5_trajsize10mdp_irl_si.tex};

\addplot [name path=upper1,draw=none] table[x=x,y expr=\thisrow{y}+\thisrow{err}]{obstacle_mem5_trajsize10mdp_irl_si.tex};
\addplot [name path=lower1,draw=none] table[x=x,y expr=\thisrow{y}-\thisrow{err}] {obstacle_mem5_trajsize10mdp_irl_si.tex};
\addplot [fill=color1!40,opacity=0.5] fill between[of=upper1 and lower1];

\addplot [name path=upper,draw=none] table[x=x,y expr=\thisrow{y}+\thisrow{err}] {obstacle_mem5_trajsize10pomdp_irl_si.tex};
\addplot [name path=lower,draw=none] table[x=x,y expr=\thisrow{y}-\thisrow{err}] {obstacle_mem5_trajsize10pomdp_irl_si.tex};
\addplot [fill=color0!40,opacity=0.5] fill between[of=upper and lower];

\end{axis}

\end{tikzpicture}
    \end{subfigure}%
    \begin{subfigure}[t]{0.541\textwidth}
        \centering
        \centering\captionsetup{width=.95\linewidth}%
\begin{tikzpicture}

\definecolor{color1}{HTML}{FF0000}
\definecolor{color0}{HTML}{0000FF}
\begin{axis}[
legend cell align={left},
legend columns=2,
legend style={
  fill opacity=0.8,
  draw opacity=1,
  text opacity=1,
  at={(0.5,1)},
  anchor=south,
  ticklabel style = {font=\tiny},
  draw=white!80!black,
  font=\fontsize{6}{6}\selectfont,
  scale=0.40
},
title={$R_\policy^\phi$},
title style={yshift=-1.5ex,xshift=-15ex,font=\titlesize},
xshift=0.1cm,
yshift = 0.5cm,
width=6.cm,
height=3.1cm,
tick align=outside,
tick pos=left,
no markers,
every axis plot/.append style={line width=2pt},
x grid style={white!69.0196078431373!black},
xlabel={\titlesize{Time Steps}},
xmajorgrids,
xmin=-4.95, xmax=105,
xtick style={color=black},
y grid style={white!69.0196078431373!black},
ylabel={\footnotesize{Mean  accumulated  reward}},
ylabel={},
ymajorgrids,
xtick={0,25,50,75,100},
ymin=-200.8979836098955, ymax=550,
ytick style={color=black}
]

\addplot[color0] table[x=x,y=y,mark=none] {obstacle_mem1_trajsize10pomdp_irl_si.tex};
\addplot[color1,dashed] table[x=x,y=y,mark=none] {obstacle_mem1_trajsize10mdp_irl_si.tex};

\addplot [name path=upper1,draw=none] table[x=x,y expr=\thisrow{y}+\thisrow{err}]{obstacle_mem1_trajsize10mdp_irl_si.tex};
\addplot [name path=lower1,draw=none] table[x=x,y expr=\thisrow{y}-\thisrow{err}] {obstacle_mem1_trajsize10mdp_irl_si.tex};
\addplot [fill=color1!40,opacity=0.5] fill between[of=upper1 and lower1];

\addplot [name path=upper,draw=none] table[x=x,y expr=\thisrow{y}+\thisrow{err}] {obstacle_mem1_trajsize10pomdp_irl_si.tex};
\addplot [name path=lower,draw=none] table[x=x,y expr=\thisrow{y}-\thisrow{err}] {obstacle_mem1_trajsize10pomdp_irl_si.tex};
\addplot [fill=color0!40,opacity=0.5] fill between[of=upper and lower];
\end{axis}

\end{tikzpicture}
    \end{subfigure}%
    \vspace*{-0.47cm}
    \caption{Representative results on the $\mathrm{Obstacle}$ example showing the reward of the policies under the true reward function ($R_\policy^\phi$) versus the time steps.}
    \label{fig:obstacle}
\end{figure}%

Similar to the $\mathrm{Maze}$ and $\mathrm{Rock}$ examples,
Figure~\ref{fig:obstacle} supports our claim that side information alleviates the information asymmetry and memory leads to more performant policies.

\clearpage
\paragraph{\textbf{Evade Instance}} 
$\mathrm{Evade}[n,r,slip]$ is a turn-based game where the agent must reach a destination without being intercepted by a faster player. 
The player cannot access the top row of the grid. 
Further, the agent can only observe the player if it is within a fixed radius from its current location and upon calling the action \emph{scan}. The parameters $n$, $r$, and $slip$ specify the dimension of the grid, the view radius, and the slippery probability, respectively.

The feature functions are defined such that the agent receives a positive reward if at the destination, a high negative reward if it is intercepted by the player, and a small negative reward for each action taken, including the \emph{scan} action.

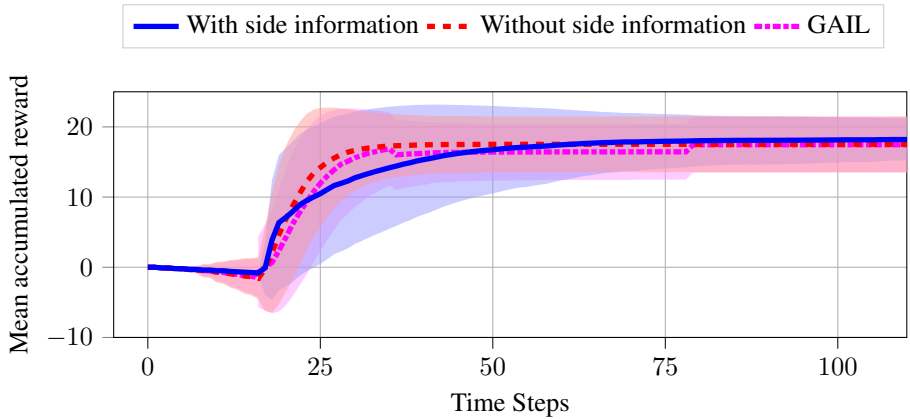
\begin{figure}[!hbt]
\centering
    \begin{subfigure}[t]{\textwidth}
        \centering\hspace*{1.4cm}\begin{tikzpicture}
\definecolor{color1}{HTML}{FF0000}
\definecolor{color0}{HTML}{0000FF}
\definecolor{color2}{HTML}{FF00FF}
\begin{customlegend}[legend columns=3,legend style={
  fill opacity=0.8,
  draw opacity=1,
  text opacity=1,
  at={(0.0,110.2)},
  anchor=south,
  draw=white!80!black,
  scale=0.40,
  mark options={scale=0.5},
},legend entries={With side information, Without side information, GAIL}]
\addlegendimage{line width=2pt, color0}
\addlegendimage{line width=2pt, color1,dashed}
\addlegendimage{line width=2pt, color2, densely dashdotted}
\end{customlegend}
\end{tikzpicture}
    \end{subfigure}\hfill%
    \vspace*{0.20cm}
    \centering
    \begin{subfigure}[t]{\textwidth}
        \centering
        \centering\captionsetup{width=.95\linewidth}%
\begin{tikzpicture}

\definecolor{color0}{HTML}{FF0000}
\definecolor{color1}{HTML}{0000FF}
\definecolor{color2}{HTML}{FF00FF}
\begin{axis}[
legend cell align={left},
legend columns=2,
legend style={
  fill opacity=0.8,
  draw opacity=1,
  text opacity=1,
  at={(0.5,1)},
  anchor=south,
  draw=white!80!black,
  font=\fontsize{6}{6}\selectfont,
  scale=0.40
},
title={$R_\policy^\phi$},
title={},
title style={yshift=-1.5ex,xshift=-15ex,font=\titlesize},
width=\textwidth,
height=0.4\textwidth,
tick align=outside,
tick pos=left,
no markers,
every axis plot/.append style={line width=2pt},
x grid style={white!69.0196078431373!black},
xlabel={{Time Steps}},
xmajorgrids,
xmin=-4.95, xmax=110,
xtick style={color=black},
y grid style={white!69.0196078431373!black},
ylabel={{Mean  accumulated  reward}},
ymajorgrids,
xtick={0,25,50,75,100},
ymin=-10, ymax=25,
ticklabel style = {font=\normalsize}
]

\addplot[color2,densely dashdotted] table[x=x,y=y,mark=none] {evade_mdp_fwd_gail.tex};
\addplot [name path=upper2,draw=none] table[x=x,y expr=\thisrow{y}+\thisrow{err}] {evade_mdp_fwd_gail.tex};
\addplot [name path=lower2,draw=none] table[x=x,y expr=\thisrow{y}-\thisrow{err}] {evade_mdp_fwd_gail.tex};
\addplot [fill=color2!40,opacity=0.5] fill between[of=upper2 and lower2];

\addplot[color0,dashed] table[x=x,y=y,mark=none] {evade_mem1_trajsize10mdp_irl.tex};
\addplot[color1] table[x=x,y=y,mark=none] {evade_mem1_trajsize10mdp_irl_si.tex};

\addplot [name path=upper1,draw=none] table[x=x,y expr=\thisrow{y}+\thisrow{err}]{evade_mem1_trajsize10mdp_irl_si.tex};
\addplot [name path=lower1,draw=none] table[x=x,y expr=\thisrow{y}-\thisrow{err}] {evade_mem1_trajsize10mdp_irl_si.tex};
\addplot [fill=color1!40,opacity=0.5] fill between[of=upper1 and lower1];

\addplot [name path=upper,draw=none] table[x=x,y expr=\thisrow{y}+\thisrow{err}] {evade_mem1_trajsize10mdp_irl.tex};
\addplot [name path=lower,draw=none] table[x=x,y expr=\thisrow{y}-\thisrow{err}] {evade_mem1_trajsize10mdp_irl.tex};
\addplot [fill=color0!40,opacity=0.5] fill between[of=upper and lower];
\end{axis}

\end{tikzpicture}
    \end{subfigure}%
    \caption{Representative results on the $\mathrm{Evade}$ example showing the reward of the policies under the true reward function ($R_\policy^\phi$) versus the time steps.}
    \label{fig:evade}
\end{figure}

As for the side information, we specify in temporal logic that while learning the reward, \emph{the agent must reach an exit point with probability greater than $0.98$.} 

Figure~\ref{fig:evade} shows that learning with side information provides higher reward than without side information. Besides, there is less randomness in the policy with side information compared to the policy without side information. Specifically, the standard deviation of the policy with side information is significantly smaller than the policy without side information.  

We did not discuss the impact of different memory size policies in this example since the performance of the memoryless policy is already near-optimal, as the policy obtains the same reward as $\texttt{SARSOP}$ (see Table~\ref{tab:ToolComp} for a reference.
Specifically, we observe that the optimal policy on the underlying MDP yields comparable policies to the optimal memoryless policy on the POMDP. 
As a consequence, we observe that the information asymmetry between the expert and the agent does not hold here either, and the learned policies obtain a similar performance.

\clearpage
\paragraph{\textbf{Intercept Instance}} 
$\mathrm{Intercept}[n,r,slip]$ is a variant of $\mathrm{Evade}$ where the agent must intercept another player who is trying to exit the gridworld. 
The agent can move in $8$ directions and can only observe the player if it is within a fixed radius from the agent's current position when the action \emph{scan} is performed. 
Besides, the agent has a camera that enables it to observe all cells from west to east from the center of the gridworld. 
In contrast, the player can only move in $4$ directions. The parameters $n$, $r$, and $slip$ specify the dimension of the grid, the view radius, and the slippery probability, respectively.

We consider three feature functions to parameterize the unknown reward. The first feature provides a positive reward to the agent upon intercepting the player. 
The second feature penalizes the agent if the player exits the gridworld. 
The third feature imposes a penalty cost for each action taken.

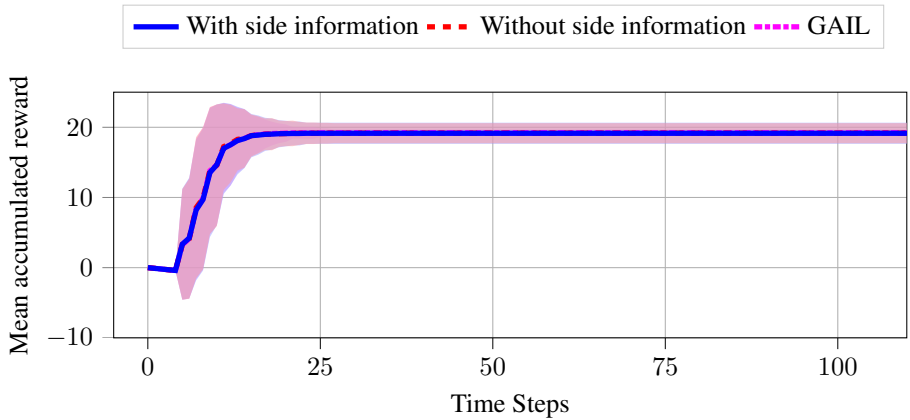
\begin{figure}[!hbt]
\centering
    \begin{subfigure}[t]{\textwidth}
        \centering\hspace*{1.4cm}\begin{tikzpicture}
\definecolor{color1}{HTML}{FF0000}
\definecolor{color0}{HTML}{0000FF}
\definecolor{color2}{HTML}{FF00FF}
\begin{customlegend}[legend columns=3,legend style={
  fill opacity=0.8,
  draw opacity=1,
  text opacity=1,
  at={(0.0,110.2)},
  anchor=south,
  draw=white!80!black,
  scale=0.40,
  mark options={scale=0.5},
},legend entries={With side information, Without side information, GAIL}]
\addlegendimage{line width=2pt, color0}
\addlegendimage{line width=2pt, color1,dashed}
\addlegendimage{line width=2pt, color2, densely dashdotted}
\end{customlegend}
\end{tikzpicture}
    \end{subfigure}\hfill%
    \vspace*{0.20cm}
    \centering
    \begin{subfigure}[t]{\textwidth}
        \centering
        \centering\captionsetup{width=.95\linewidth}%
\begin{tikzpicture}

\definecolor{color0}{HTML}{FF0000}
\definecolor{color1}{HTML}{0000FF}
\definecolor{color2}{HTML}{FF00FF}
\begin{axis}[
legend cell align={left},
legend columns=2,
legend style={
  fill opacity=0.8,
  draw opacity=1,
  text opacity=1,
  at={(0.5,1)},
  anchor=south,
  draw=white!80!black,
  font=\fontsize{6}{6}\selectfont,
  scale=0.40
},
title={$R_\policy^\phi$},
title={},
title style={yshift=-1.5ex,xshift=-15ex,font=\titlesize},
width=\textwidth,
height=0.4\textwidth,
tick align=outside,
tick pos=left,
no markers,
every axis plot/.append style={line width=2pt},
x grid style={white!69.0196078431373!black},
xlabel={{Time Steps}},
xmajorgrids,
xmin=-4.95, xmax=110,
xtick style={color=black},
y grid style={white!69.0196078431373!black},
ylabel={{Mean  accumulated  reward}},
ymajorgrids,
xtick={0,25,50,75,100},
ymin=-10, ymax=25,
ticklabel style = {font=\normalsize}
]

\addplot[color2,densely dashdotted] table[x=x,y=y,mark=none] {intercept_mdp_fwd_gail.tex};
\addplot [name path=upper2,draw=none] table[x=x,y expr=\thisrow{y}+\thisrow{err}] {intercept_mdp_fwd_gail.tex};
\addplot [name path=lower2,draw=none] table[x=x,y expr=\thisrow{y}-\thisrow{err}] {intercept_mdp_fwd_gail.tex};
\addplot [fill=color2!40,opacity=0.5] fill between[of=upper2 and lower2];

\addplot[color0,dashed] table[x=x,y=y,mark=none] {intercept_mem1_trajsize10mdp_irl.tex};
\addplot[color1] table[x=x,y=y,mark=none] {intercept_mem1_trajsize10mdp_irl_si.tex};

\addplot [name path=upper1,draw=none] table[x=x,y expr=\thisrow{y}+\thisrow{err}]{intercept_mem1_trajsize10mdp_irl_si.tex};
\addplot [name path=lower1,draw=none] table[x=x,y expr=\thisrow{y}-\thisrow{err}] {intercept_mem1_trajsize10mdp_irl_si.tex};
\addplot [fill=color1!40,opacity=0.5] fill between[of=upper1 and lower1];

\addplot [name path=upper,draw=none] table[x=x,y expr=\thisrow{y}+\thisrow{err}] {intercept_mem1_trajsize10mdp_irl.tex};
\addplot [name path=lower,draw=none] table[x=x,y expr=\thisrow{y}-\thisrow{err}] {intercept_mem1_trajsize10mdp_irl.tex};
\addplot [fill=color0!40,opacity=0.5] fill between[of=upper and lower];
\end{axis}

\end{tikzpicture}
    \end{subfigure}%
    \caption{Representative results on the $\mathrm{Intercept}$ example showing the reward of the policies under the true reward function ($R_\policy^\phi$) versus the time steps.}
    \label{fig:intercept}
\end{figure}

We encode the high-level side information as the temporal logic task specification \emph{Eventually intercept the player with probability greater than $0.98$}, i.e., the agent should eventually reach an observation where its location coincides with the player's location.

Figure~\ref{fig:intercept} demonstrates that side information does not improve the performance of the policy. 
This result is because memoryless policies are optimal in this example, and a combination of the given reward features can perfectly encode the temporal logic specifications, similar to the $\mathrm{Evade}$ example.

\subsection{\myett{Effects of Side Information}}
\myett{In this section, we provide additional experiments on how the side information speeds up the learning process in terms of computation time and convergence to the optimal policy and reward parameters. Then, we quantify the effects of the side information by solving the POMDPs using only the task specifications and no demonstrations. All these experiments are performed on the Maze example, which is a relatively low-size POMDP example.}

\myett{\paragraph{\textbf{Convergence of the Learning With and Without Side Information}}
In the Maze experiments, we empirically observe that side information enables the learning algorithm to converge with \emph{eight} times less number of iterations compared to learning without side information. The number of iterations here denotes both the number of linearizations in the sequential convex scheme and the number of gradient steps during reward updates. However, the gain in computation time is not as prominent as the gain in the number of iterations. In fact, learning with side information is only approximately \emph{three} times faster than learning without side information due to how side information almost doubles the number of variables in the convex optimization problem.
}

\myett{\paragraph{\textbf{Effects of Side Information Without Any Demonstrations}}
\begin{itemize}
    \item \emph{Experiment 1.} We consider the exact setting of the Maze example with no demonstrations and the LTL specification $\mathrm{Pr}_{\Pomdp}^\policy(\textbf{G} \; \lnot \: \mathrm{bad}) \geq 0.9$. Essentially, we seek policies that avoid the trapping states with high probability. Without any additional reward, the optimal policy is exactly what we expect: High probability for action enforcing no movements and low probability for the others. With respect to the true reward, this is clearly suboptimal as the reward will keep decreasing due to no minimization of the amount of time spent in the environment. Now, we optimize the problem for a policy that satisfies the LTL specifications while penalizing spending time in the environment according to the ground truth reward for taking more steps in the environment. The obtained policy has the optimal reward of $47.83$, which corresponds to optimizing the ground truth reward in the POMDP. Indeed, the fastest way to clear the Maze is to exit through a goal state while not getting trapped.
    \item \emph{Experiment 2.} We consider the exact setting of the Maze example with no demonstrations and the LTL specification $\mathrm{Pr}_{\Pomdp}^\policy(\textbf{E} \;  \mathrm{target}) \geq 0.95$. Essentially, we seek policies that eventually reach the target state (exit of the Maze) with high probability. Without any additional reward, the optimal policy is suboptimal with respect to the optimal policy on the POMDP, given the ground truth reward. Indeed, the policy can reach the target with an optimal reward of $28.63$ due to the amount of time spent to reach the goal. By adding the reward on time spent in the environment to the LTL specifications, we can obtain the optimal reward on the POMDP again.
\end{itemize}
}

\subsection{Summary of the Results}

\paragraph{\textbf{Side Information Alleviates the Information Asymmetry} }
As mentioned in the submitted manuscript, side information can indeed alleviate the information asymmetry.
Specifically, we observe that if there is an information asymmetry in the \emph{forward problem}, i.e., the obtained reward from an optimal policy on the underlying POMDP is lower than from an optimal policy on the underlying fully observable MDP, incorporating side information in temporal logic specifications alleviates the information asymmetry between the expert and the agent.
For example, we can see the effects of such information asymmetry in the $\mathrm{Maze}$, $\mathrm{Rocks}$, $\mathrm{Obstacle}$, and $\mathrm{Avoid}$ examples.
In these examples, having partial observability reduces the obtained reward in the forward problem. 
The policies that do not incorporate side information into the learning procedure also obtain a lower reward under information asymmetry.

\paragraph{\textbf{Memory Leads to More Performance Policies}}
Similarly to the side information, we also observe that if incorporating memory improves the performance of the learned policies, if it also improves the obtained reward in the forward problem, as seen in the $\mathrm{Maze}$, $\mathrm{Rocks}$, and $\mathrm{Obstacle}$ instances.
In Table~\ref{tab:ToolComp}, we can also see that incorporating memory helps to compute a better optimal policy in these examples, unlike computing a memoryless policy.

\end{appendices}

\end{document}